\documentclass[10pt,twocolumn,letterpaper]{article}

\usepackage{wacv}
\makeatletter
\@namedef{ver@everyshi.sty}{}
\makeatother

\usepackage{times}
\usepackage{epsfig}
\usepackage{graphicx}
\usepackage{amsmath}
\usepackage{amssymb}
\usepackage{booktabs}

\usepackage{algorithmic}
\usepackage{algorithm}
\usepackage{array}
\usepackage{textcomp}
\usepackage{stfloats}
\usepackage{url}
\usepackage{verbatim}
\usepackage{cite}

\usepackage{amsfonts,amssymb,amsthm}
\usepackage{framed}
\usepackage{bbm}
\usepackage[labelfont=bf]{caption}
\usepackage{subcaption}
\usepackage{float}
\usepackage{makecell}
\usepackage{longtable}
\usepackage{enumitem}
\usepackage{lipsum}
\usepackage{todonotes}
\usepackage{symbols}
\usepackage{tabularx}

\def\wacvPaperID{****} 

\wacvalgorithmstrack   

\wacvfinalcopy 

\ifwacvfinal
\usepackage[breaklinks=true,bookmarks=false]{hyperref}
\else
\usepackage[pagebackref=true,breaklinks=true,colorlinks,bookmarks=false]{hyperref}
\fi

\begin{document}

\title{Efficient Human Vision Inspired Action Recognition\\using Adaptive Spatiotemporal Sampling}

\author{
    Khoi-Nguyen C. Mac$^\dagger$, Minh N. Do$^\dagger$, Minh P. Vo$^\ddagger$\\
	$^\dagger$University of Illinois at Urbana-Champaign, $^\ddagger$Meta Reality Labs Research \\
	$^\dagger${\tt\small \{knmac, minhdo\}@illinois.edu}, $^\ddagger${\tt\small minh.vo@fb.com}
}

\maketitle
\thispagestyle{empty}

\begin{abstract}
    Adaptive sampling that exploits the spatiotemporal redundancy in videos is critical for always-on action recognition on wearable devices with limited computing and battery resources. The commonly used fixed sampling strategy is not context-aware and may under-sample the visual content, and thus adversely impacts both computation efficiency and accuracy. Inspired by the concepts of foveal vision and pre-attentive processing from the human visual perception mechanism, we introduce a novel adaptive spatiotemporal sampling scheme for efficient action recognition. Our system pre-scans the global scene context at low-resolution and decides to skip or request high-resolution features at salient regions for further processing. We validate the system on EPIC-KITCHENS and UCF-101 datasets for action recognition, and show that our proposed approach can greatly speed up inference with a tolerable loss of accuracy compared with those from state-of-the-art baselines.
    Source code is available in {\small\url{https://github.com/knmac/adaptive_spatiotemporal}}.
\end{abstract}


\section{Introduction}
\label{sec:intro}

\begin{figure}
	\centering
	\includegraphics[width=\linewidth]{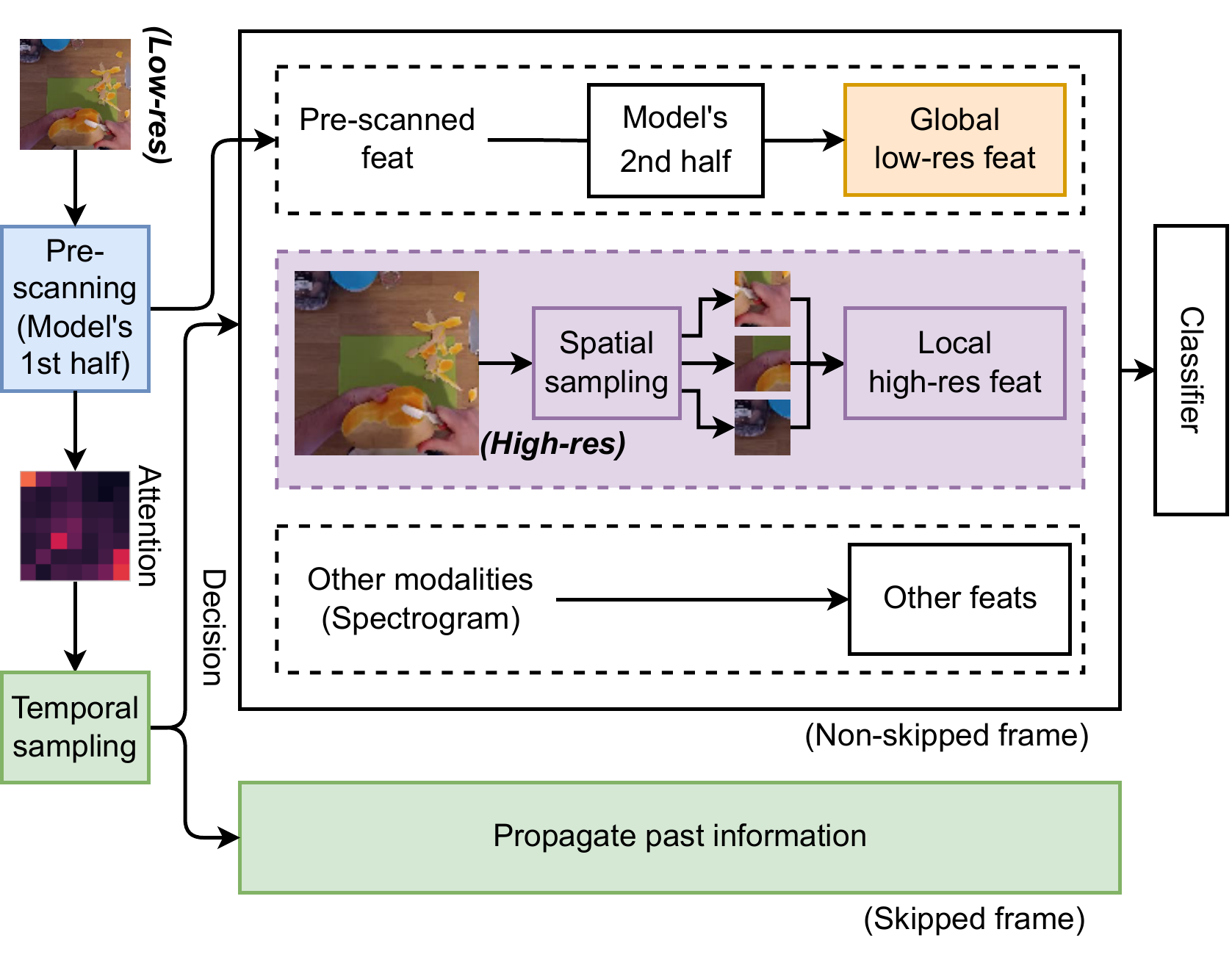}
	\caption{Our proposed system has two major components: temporal and spatial sampling. Based on a pre-scanned features, the temporal sampler decides whether to process the frame fully (Full model), or skip to the frame and propagate past information (bottom block). The spatial sampler in turns select RoIs from high-res input to augment the features with low-res inputs. We also include features from other available modalities if a frame is fully processed. We color-code the spatial sampling as purple and temporal sampling as green. We further illustrate details of the two routines in \figref{fig:space_sampler} and \figref{fig:time_sampler}.}
	\label{fig:overview}
	\vspace{-.6cm}
\end{figure}

Our visual world is highly predictive, making it highly inefficient to process each individual piece of data with the same amount of effort. To cope with it, human perceptual system subconsciously pre-scans the scene to determine important events before actual processing. This mechanism is known as \textit{pre-attentive} processing \cite{appelbaum2009attentive, atienza2001auditory, meng2009pre}. The pre-capturing images, although appear to be less clear, constructs the global perception of the scene \cite{barghout1999differences, klein1997seven}. Furthermore, the human brain also focuses on certain regions within our \textit{foveal} visual field \cite{barghout1999differences, iwasaki1986relation, kolb2011simple, kourtzi2000cortical}. These two behaviors are strikingly similar to the objective of our temporal and spatial sampling, respectively.

Sampling has also been one of the most studied problems in various areas of video analysis, such as action recognition and video summarization \cite{gong2014diverse, gygli2015video, korbar2019scsampler, mahasseni2017unsupervised, zhang2016real, zhang2016video}, due to the redundancy between consecutive frames. With the increase in model complexity, it gets progressively expensive to process a single frame. This is even more crucial for resource-limited devices such as AR/VR headsets, like Google Glasses, HoloLens, Ray-Ban Stories, \etc \cite{googleglass, hololens, rayban}. However, picking a fixed sampling scheme does not guarantee the performance as important information may be under-sampled. Temporally, it is evident that the number of frames required to represent a video vary, depending on the action categories \cite{huang2018makes}. Therefore, an adaptive sampling rate is preferred as over-sampling results in more computational cost while under-sampling can make performance suffer. Similarly, spatial sampling is also necessary in general computer vision tasks, which is applicable on individual frames of a video sequence. It is preferred to have an adaptive sampling scheme as having a fixed one also leads to similar problems as in time domain.

Inspired by the mechanism of human visual perception, we propose a novel adaptive spatiotemporal sampling framework to imitate the human vision. \figref{fig:overview} shows an overview of the entire system, with two main components that are built upon the self-attention mechanism: (1) spatial sampler uses the observed attention to sample regions of interest and (2) temporal sampler hallucinates attention in the next frame to model future expectation.

\begin{figure}
	\centering
	\includegraphics[width=\linewidth]{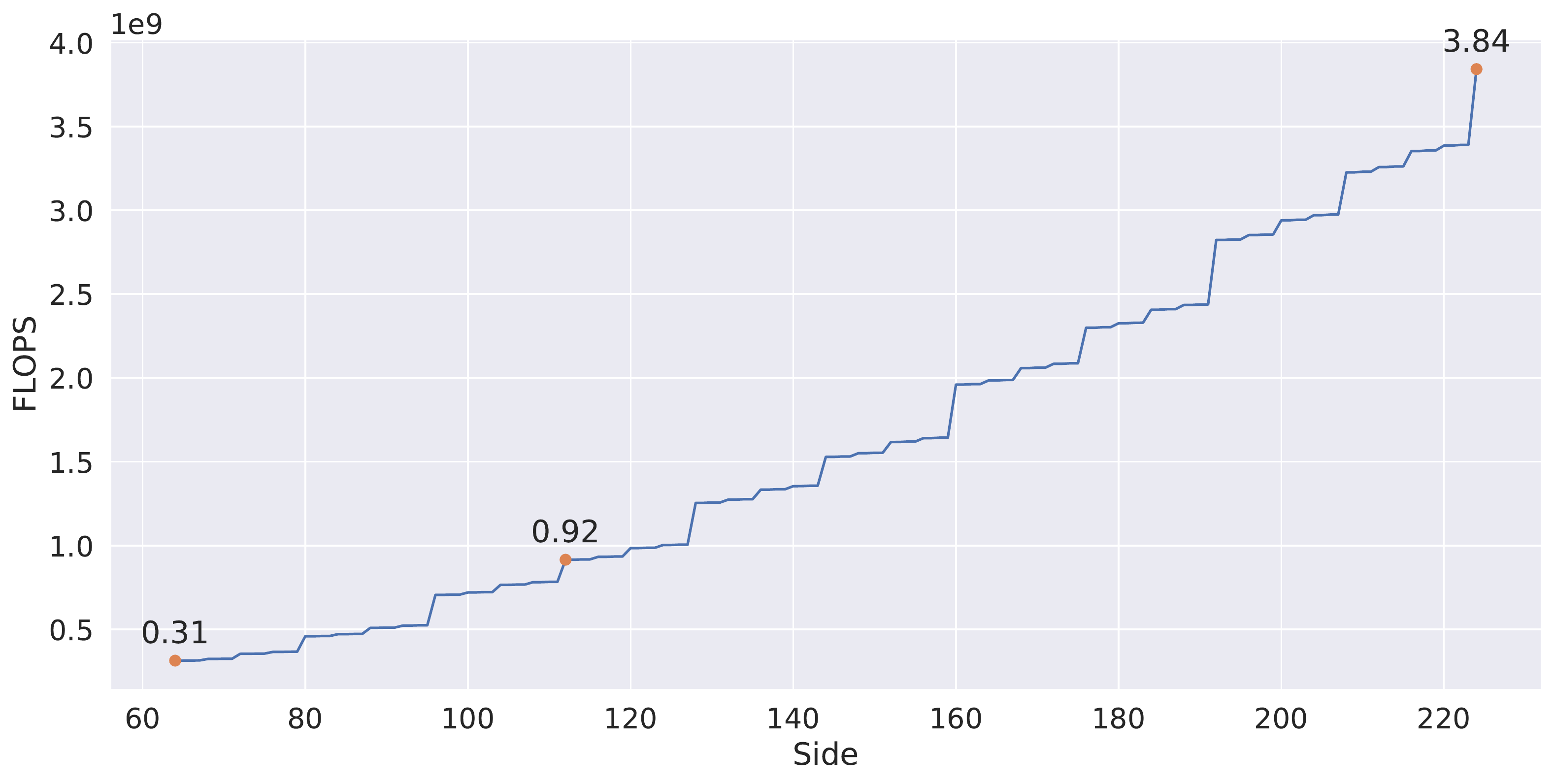}
	\caption{Complexity of SAN19 with different input dimensions. The horizontal axis shows the side $N$ of an input with dimension $3 \times N \times N$. We highlight the values where $N = 224, 112, 64$, corresponding to our choices for the size of high-res, low-res, and cropped high-res images.}
	\label{fig:resolution_analysis}
	\vspace{-.5cm}
\end{figure}

The \textit{spatial sampler} is motivated by human \textit{foveal vision}. The idea is to only focus on specific regions rather than the whole scene to save computation. It can be seen that lower input sizes significantly reduce the computational complexity. However, it also compromises the performance. To overcome this, we use input at two different resolutions: low-res whole image with size of 112x112 and high-res image crops with size of 64x64. The low-res images are processed as a whole for pre-scanning process of temporal sampler and global feature extraction. For the high-res input, we retrieve regions corresponding to the most ``important'' locations based on the extracted attention and use them to augment the low-res image for the visual recognition task. In our system, we use the low-res image of size 112x112 and top-$k$ regions of size 64x64. \figref{fig:resolution_analysis} analyzes the computational complexity in GFLOPS with respect to the spatial dimension of RGB images. We highlight the GFLOPS with size 224x224 (high-res baseline), compared with size 112x112 (low-res input) and size 64x64 (cropped high-res regions).

The \textit{temporal sampler} follows the concept of \textit{pre-attentive processing} such that it extracts attention by briefly pre-scanning a low-res input and decides whether to further process the frame if something interesting happens. Since it is possible to predict what would happen in the future \cite{gao2018im2flow, guan2020generative}, we consider an event ``interesting'' if it is drastically different from what is expected. The idea of using another network for pre-scanning has been discussed in other work \cite{korbar2019scsampler, meng2020arnet}, however in our proposed approach, we split a backbone network into two halves and use the first one to pre-scan instead of introducing an additional one. We observe that two consecutive frames produce similar attention at some certain intermediate layers, making it possible to pre-scan by forwarding up to such layers. To model future expectation, we hallucinate future attention and compare it with the observed one. When the hallucination matches the actual attention, there is no unexpected event and the model simply uses the previous classification results. Otherwise, the remaining processing routine is carried out to compute new classification. For further analogy about human visual perception, please refer to the supplementary materials.

We demonstrate the effectiveness of our system on the action recognition task on EPIC-KITCHENS \cite{damen2018epickitchens} and UCF-101 \cite{soomro2012ucf101} datasets. Our system reduces computational complexity with a tolerable loss of accuracy compared to the baseline counterparts. We also provide qualitative results to reason the sampling results.

\noindent\textbf{Contribution:} (1) We introduce a novel adaptive spatiotemporal sampling scheme, where the temporal sampler can pre-scan the low-res input to decide whether to skip processing by comparing the observed and hallucinated attention. (2) As a part of the sampling routine, our spatial sampler selects small high-res RoIs induced by the attention map in pre-scanning process. (3) We showcase the system on egocentric and generic videos where our model reduces the computational power with a small loss of accuracy.


\section{Related work}
\label{sec:relatedwork}

\noindent\textbf{Action recognition.} With the blooming of deep learning and computer vision, the task of action recognition has evolved from the traditional two-stream networks \cite{simonyan2014twostream} to more advanced models, \eg, C3D, I3D, ResNet3D, R(2+1)D, TBN, TSN, and LSTA \cite{tran2015c3d, carreira2017i3d, hara2018resnet3d, tran2018r2+1d, kazakos2019tbn, sudhakaran2019lsta, wang2016tsn}. Such standard techniques often demand expensive computation, leading to the challenge of high power consumption \cite{possas2018egocentric}, which is crucial for action recognition using always-on wearable devices, such as AR/VR glasses. Our adaptive sampling scheme aims to address this problem.

\noindent\textbf{Adaptive inference and sampling.} Techniques to reduce the complexity of deep networks can be divided into three sub-categories: ignoring layers in deep models, removing input regions, and skipping frames. \cite{huang2016deep} introduces a stochastic method to drop layers during the training phase. SkipNet \cite{wang2018skipnet} and BlockDrop \cite{wu2018blockdrop} later propose to use reinforcement learning to dynamically drop layers for both training and validation. In spatial domain, RS-Net \cite{wang2020resolution} can decide which resolution to switch to by sharing parameters among different image scales. PatchDrop \cite{uzkent2020learning}, on the other hand, removes unimportant regions of input images via reinforcement learning. For applications in the area of general video analysis, it is more desirable to rely on time sampling. It has been shown that temporal redundancy results in wasted computation, as some videos only require a single frame to represent \cite{huang2018makes}. There have been attempts to process videos at multiple frame rates as different actions can happen at different paces \cite{feichtenhofer2019slowfast, xiao2020audiovisual}. Recently, SC-Sampler \cite{korbar2019scsampler} and ARNet \cite{meng2020arnet} tackle temporal sampling by using additional simple networks for pre-scanning the features. \cite{wang2021adaptive} focuses on spatial redundancy and frame-skipping is a special case in their formulation. In contrast, \cite{kim2021efficient} focuses on the time domain and do not consider partial spatial information. We explicitly model spatiotemporal sampling using self-attention, as inspired by human vision, and only use a sub-collection of layers to pre-scan. Our model is also more interpretable with visualizable attention and hallucination.

\begin{figure}
	\centering
	\includegraphics[width=\linewidth]{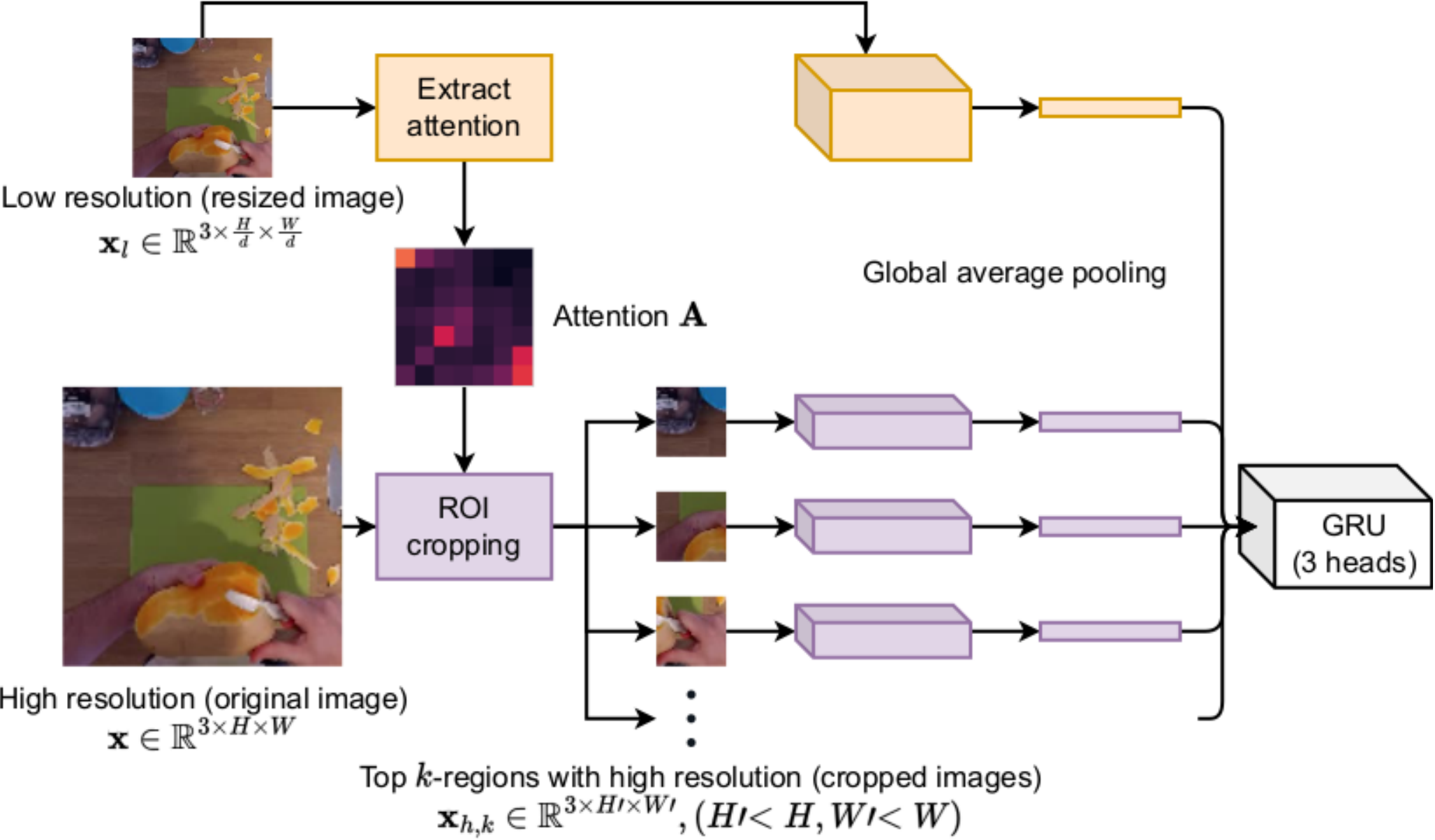}
	\caption{Spatial sampler uses attention from low-res image to sample the top-$k$ regions from the (original) high-res input. $\bx_l$ gives a global view, while $\bx_{h,k}$ provides local views at important regions of the original image $\bx$. The final global average pooling removes spatial dimension of the features, which are combined and fed to the three-head GRU classifier. The heads correspond to low-res features, high-res features, and their concatenation, and are used to encourage strong learning feature at each resolution.}
	\label{fig:space_sampler}
	\vspace{-.5cm}
\end{figure}

\noindent\textbf{Self-attention.} In computer vision, gradient-based methods are usually used to generate saliency maps, which can determine the regions where a trained model considers ``relevant'' to the output \cite{julius2018sanity_check, ribeiro2016lime, selvaraju2017gradcam, zeiler2014visualizing}. More recently, self-attention is introduced in natural language processing community as a way to direct the focus of deep nets \cite{vaswani2017attention}. Since the attention allows a model to focus more on important regions, such self-attention mechanism has been attracting great interest from the computer vision community \cite{zhao2020san, ramachandran2019standalone, ren2020psam}. Our approach uses such attention as a driving mechanism to find the important regions and frames, allowing spatiotemporal sampling adaptively. More recently, vision transformer has been introduced as another attention-based approach and adopted in several work \cite{dosovitskiy2021vit, arnab2021vivit, yin2022avit, meng2022adavit, wu2022memvit, li2022mvitv2}. Since our method operates on top of attention, the backbone models can be flexibly interchanged with any attention-based CNNs. Here, we choose to use the SAN-19 backbone \cite{zhao2020san} and focus on improving efficiency of the baseline models.

\section{Approach}
\label{sec:approach}

Consider a video dataset $\mathcal{D} = \{(\bv_n, \by_n)\}_{n=1}^N$, where $\bv_n$ is a video sequence and $\by_n$ is the corresponding groundtruth label. We assume that the video sequences have the same length of $T$ frames, \ie, $\bv_n = [\bx_n^{(1)}, \bx_n^{(2)}, \dots, \bx_n^{(T)}]$, where each frame is $\bx_n^{(t)} \in \R^{3 \times H \times W}, t \in \{1, .., T\}$. Suppose that we have a video classifier $F(\bv_n) = \hat{\by}_n$, with some complexity $\mathcal{O}_F$. The goal is to construct another classifier $\tilde{F}$ with less complexity while retaining the accuracy. We address this by introducing the temporal sampler $\mathcal{T}$ and spatial sampler $\mathcal{S}$, \ie, $\hat{\by}_n = \tilde{F}(\bx_n; \mathcal{T}, \mathcal{S})$, such that $\mathcal{O}_{\tilde{F}} < \mathcal{O}_F$. At a high level, the spatial sampler chooses the top-$k$ regions based on the most activated areas in the attention map. The temporal sampler decides whether to skip frames, whose attentions are similar to the model's future prediction.

\subsection{Cumulative global attention}
\label{ssec:global_attention}

\begin{figure}
    \centering
    \includegraphics[trim=69em 0 0 0, clip, width=0.77\linewidth]{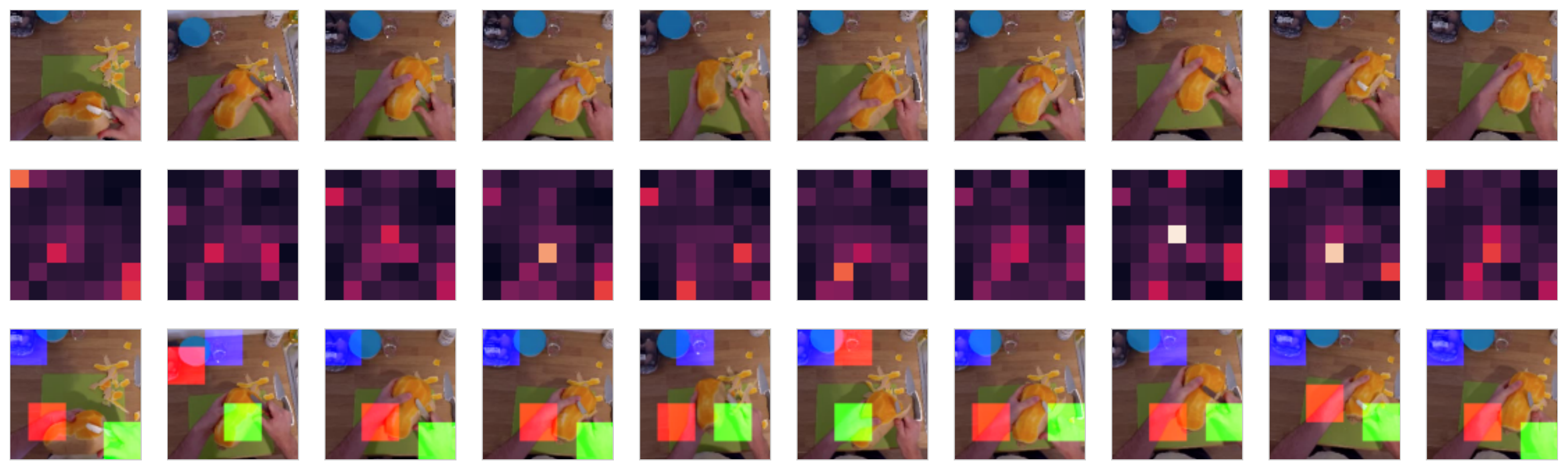}
    \caption{Sampled regions from the top-3 spatial sampler. From top to bottom: (1) input frames, (2) attention, and (3) bounding boxes in pixel space. Red, green, and blue colors denote the top 1, 2, and 3 accordingly. The trajectories of the boxes reflect to their activated regions. Such behavior allows prediction of future attentions.}
    \label{fig:attn_bbox}
    \vspace{-.5cm}
\end{figure}

Our cumulative global attention is built upon the pairwise attention formulation of Zhao et al. \cite{zhao2020san}. We rewrite this pairwise attention as
\begin{equation}
    \bz_i = \sum_{j \in \mathcal{R}(i)} \alpha\left( Q(\bx_i), K(\bx_j) \right) \odot V(\bx_j),
\end{equation}
where $i, j \in \R^2$ are the spatial indices, $Q(\bx_i)$, $K(\bx_j)$, and $V(\bx_j)$ are the query, key, and value encodings, and $\alpha$ is the compatibility function, usually defined as a softmax. Such compatibility function is locally defined over the footprint $\mathcal{R}(i)$. We then denote the \textit{local attention} at $i$ as
\begin{equation}
    \ba_i = \left[ \alpha\left( Q(\bx_i), K(\bx_j) \right) \right], \quad j \in \mathcal{R}(i).
    \label{equ:local_attention}
\end{equation}
Learning to generate such attentions is difficult because we also need to model the underlying relationship of neighboring footprints, \ie, $a_i$ and $a_{i+1}$ have overlapping footprint. However, it is simpler to generate a global attention map where the footprints are already encoded. Thus, we use the \textit{cumulative global attention}, defined as
\begin{equation}
    \bA = \sum_i \ba_i \otimes \mathbbm{1}\{\mathcal{R}(i)\},
    \label{equ:global_attention}
\end{equation}
where $\mathbbm{1}\{\mathcal{R}(i)\}$ is the indicator function that removes locations outside of footprint $\mathcal{R}(i)$ and $\otimes$ is the multiplication of $\ba_i$ with the corresponding footprint. Notice that $\ba_i$ has the same spatial dimension as $\mathcal{R}(i)$, while $\bA$ has the same spatial dimension as the input feature map $\phi(\bx)$. For simplicity, unless stated otherwise we use ``attention'' to denote the cumulative global attention in the remaining sections. Please refer to our supplementary materials for further analysis on the local and global attention maps.

\subsection{Spatial sampler}
\label{ssec:spatial_sampler}

\begin{figure*}
    \centering
    \includegraphics[width=0.85\linewidth]{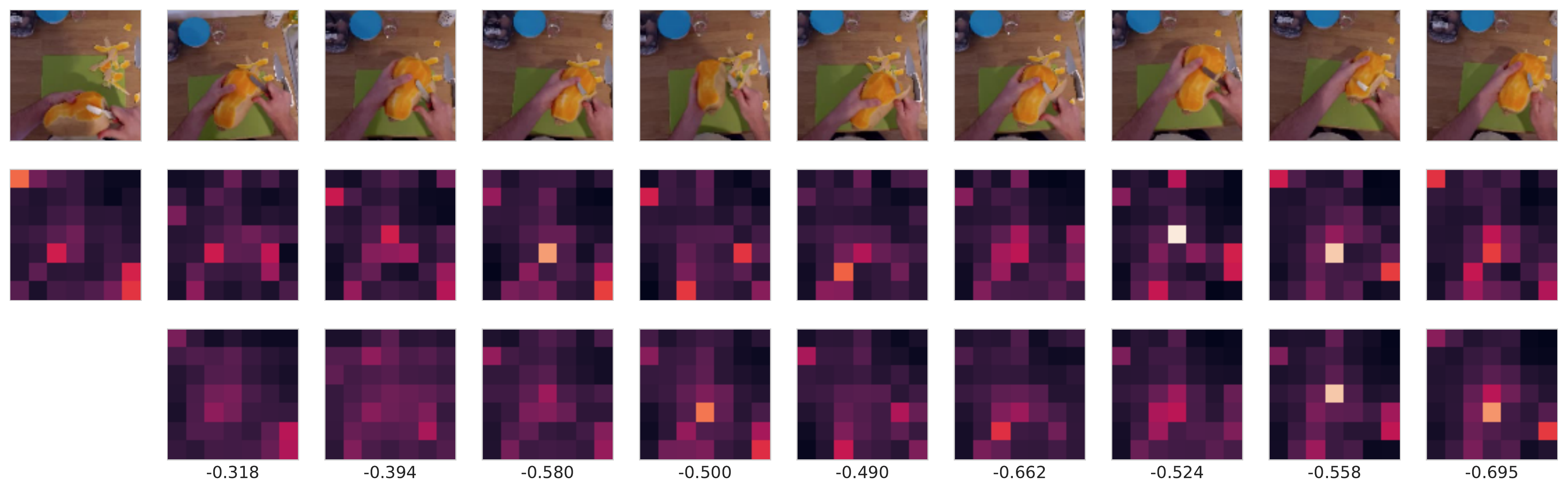}
    \caption{Attention and corresponding hallucination of a video sequence. From top to bottom: (1) input frames, (2) attention, and (3) hallucination. Negative SSIM scores between the attention and hallucination are included at the bottom (0 means most different and -1 means most similar). Activated regions of the attentions and hallucinations match the movements of the hands across frames, showing the temporal consistency property.}
    \label{fig:attn_hallu}
    \vspace{-.5cm}
\end{figure*}

The goal of the spatial sampler is to provide the high-resolution inputs at locations where it matters, which is similar to foveal vision in human.
Formally, given input $\bx \in \R^{3 \times H \times W}$, we compute the corresponding low and high-res inputs $\bx_l \in \R^{3 \times \frac{H}{d} \times \frac{W}{d}}$ and $\bx_{h,k} \in \R^{3 \times H' \times W'}$, respectively obtained by rescaling (with the down-sampling factor $d$) and cropping $\bx$ ($H' < H, W' < W$) at $k$ different locations. While $d$ is defined as our hyper-parameter, the cropping regions for $\bx_{h,k}$ are computed by the spatial sampler $\mathcal{S}$.
Giving attention $\bA$, we find all connected regions and pick the $k$ regions with highest summation. We then linearly project those regions back to pixel space, based on the scaling of spatial dimension between the input image $\bx$ and the attention $\bA$.

\figref{fig:space_sampler} shows the details of the spatial sampler. We extract the attention from the low-res image $\bx_l$ and use it to sample the top-$k$ regions in the original image $\bx$. This results in $\bx_{h,k}$ with lower spatial dimension, while retaining the original resolution of $\bx$. As we use the same backbone network to process images of different resolution, we add a global average pooling layer at the end of the feature extractor to remove the spatial dimension. The features are then fed to the three-head GRU classifier. The heads correspond to low-res features, high-res features, and their concatenation and are used to encourage strong learning feature at each resolution. We constraint the scaling factor $d$ and the bounding box size $H', W'$ such that the complexity of using $\bx_l$ and $\bx_{h,k}$'s is less than that of $\bx$. We choose $d=2$ and $H' = W' = 64$, based on our complexity analysis in \figref{fig:resolution_analysis}.

In \figref{fig:attn_bbox}, we illustrate some example results of our spatial sampler, extracting the top 3 regions in a few frames of a video sequence. The colors here denote the order of the bounding boxes, based on the most activated regions in the attention. It is observed that the sampled regions are not varying rapidly when the activation are similar. This usually happens when the actions are occurring slowly, suggesting that we can predict future attentions in such cases.

\subsection{Hallucinator}
\label{ssec:hallucinator}

Our hallucinator $H$ is grounded on the notation of temporal consistency, \ie, the attentions of consecutive frames $\bA^{(t)}$ and $\bA^{(t+1)}$ are similar if the corresponding actions  are close. Intuitively, the attention can reflect the important regions of input images and is not expected to change drastically between consecutive frames. We use $H$ to predict future attention, from which the model can decide to skip future frames if the prediction matches the observed pre-scanned features. The hallucinator $H$ is written as
\begin{equation}
    \tilde{\bA}^{(t+1)} = H ( \bA^{(t)} ) \quad \text{s.t.} \quad \tilde{\bA}^{(t+1)} \approx \bA^{(t+1)},
\end{equation}
where $\tilde{\bA}^{(t+1)}$ is the hallucination (predicted future attention). To quantify the similarity between $\tilde{\bA}^{(t+1)}$ and $\bA^{(t+1)}$, we use the structural similarity index measure (SSIM) \cite{wang2003multiscale} as this metrics can compare the structure of input tensors. We train the hallucinator by minimizing our belief loss:
\begin{equation}
    \mathcal{L}_b = -\frac{1}{T-1} \sum_{t=2}^{T} \text{SSIM} ( H(\bA^{(t-1)}), \bA^{(t)} ),
    \label{equ:belief_loss}
\end{equation}
where the function $\text{SSIM}()$ computes the structural similarity between the hallucination $H(\bA^{(t-1)})$ and the attention $\bA^{(t)}$. We minimize the negative SSIM score since the default SSIM ranges from 0 to 1. Higher SSIM indicates more similarity. We build the hallucinator as a convolutional LSTM \cite{shi2015conlstm} with encoder-decoder layers and apply teacher forcing technique \cite{williams1989teacherforcing} for the training routine.

\figref{fig:attn_hallu} shows an example of the hallucination from a video sequence, where the first row is the input video sequence, the second row is the attention extracted from a layer, and the last row is the hallucination, generated by our hallucinator. There is a missing hallucination at the first frame because we are generating future attention. It is observed that the most activated regions of the attention here are located around the two hands. As the hands move in time, these regions also move with a similar manner, in both the attention and hallucination. It suggests that our hallucinator can predict where the important regions would be in the future. We also provide the negative SSIM scores at the bottom to compare the structural similarity between the attention and hallucination. Note that our objective here is not to generate a perfect hallucination, but only to use it as a guideline for the temporal sampler.

\subsection{Temporal sampler}
\label{ssec:temporal_sampler}

Given a video sequence $\bv = [\bx^{(1)}, \dots, \bx^{(T)}]$, the objective of the temporal sampler is to adaptively select a subset of ``important'' frames that can still represent $\bv$, similar to human's pre-attentive processing mechanism. A frame $\bx^{(t)}$ is considered as unimportant if we can reasonably predict its the attention. From \secref{ssec:global_attention}, we can retrieve the attention and hallucination at any arbitrary layer from a model of $L$ layers. Suppose that the attention is extracted at layer $\lambda < L$, it is wasteful to compute the last $L-\lambda$ layers if the temporal sampler decides to skip this frame. In other words, we can forward a frame up to layer $\lambda$ and choose to run the rest of the model adaptively.

\begin{figure}
	\centering
	\includegraphics[width=\linewidth]{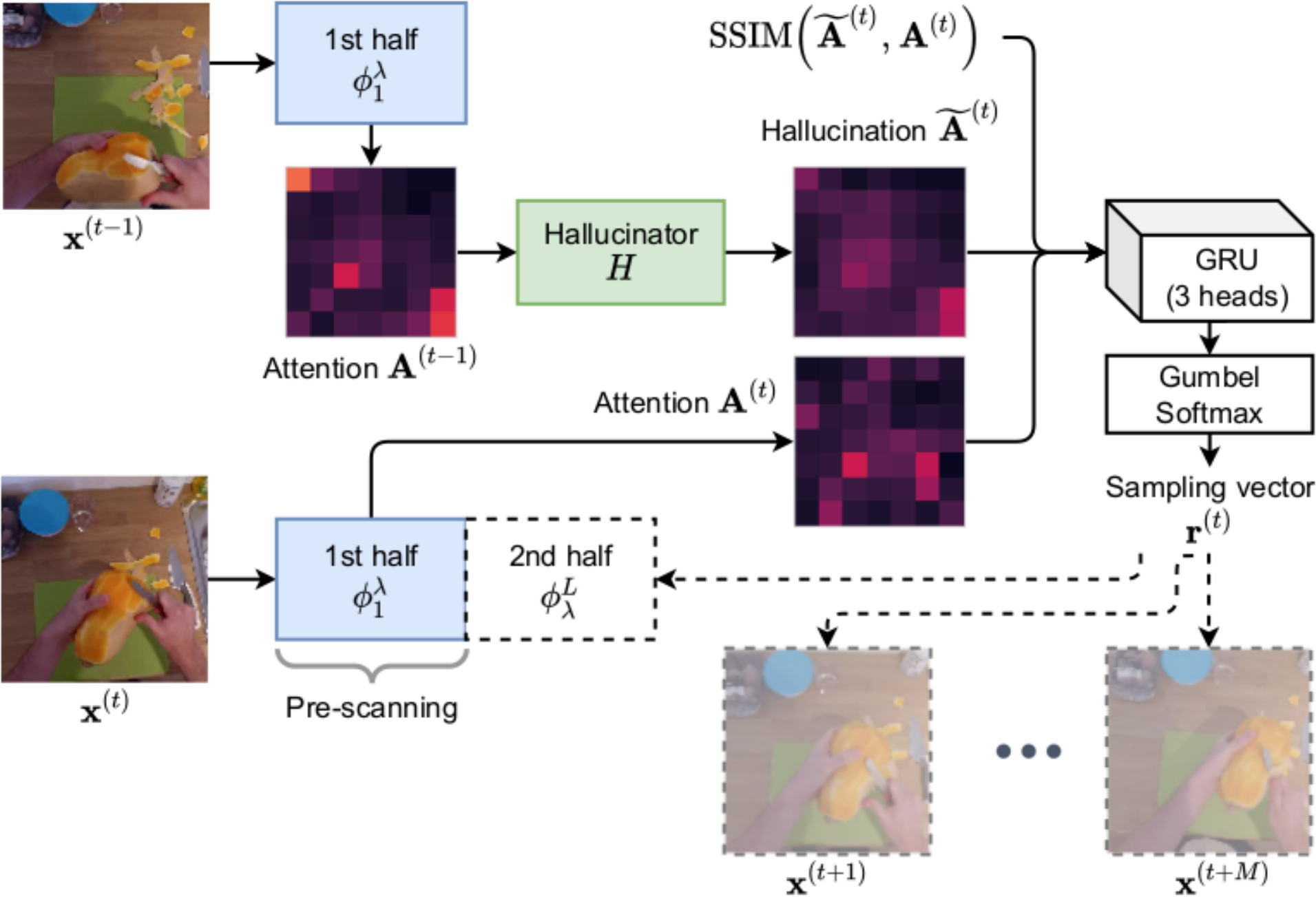}
	\caption{Temporal sampler with inputs at $t-1$ and $t$. Attention from the model's first half at time $t$, hallucination computed at time $t-1$, and their SSIM score are fed to a GRU to compute the sampling vector $\br^{(t)}$, deciding how many frames to skip (including the second half of the current frame). Model weights are shared across frames.}
	\label{fig:time_sampler}
	\vspace{-.5cm}
\end{figure}

Formally, consider the feature extractor of a deep network of $L$ layers as a composite function, we can split it into two halves at layer $\lambda \in \{1, ..., L\}$, \ie, $\phi_1^L(\bx) = \phi_\lambda^L (\phi_1^\lambda (\bx))$. The first half $\phi_1^\lambda$ is used for pre-scanning while the second half $\phi_\lambda^L$ can also be augmented with information from other modalities for the classification task later. The temporal sampler $\mathcal{T}$ determines the sampling routine by computing a sampling vector $\br = [\br^{(1)}, \dots, \br^{(T)}]$, \ie, $\mathcal{T}(\bv) = [\bx^{(t)} \times \br^{(t)}]_{t=1}^{T}$, with $\br^{(t)} \in \{0, 1\}^{M+1}$, where $\br^{(t)}[m]=1$ means we can skip $m$ frames, $m \in \{0, ..., M\}$. \figref{fig:time_sampler} shows the details of the temporal sampler. At time $t$, the attention $\bA^{(t)}$ is extracted using the first half of the feature extractor $\phi_1^\lambda$. To generate the sampling vector $\br^{(t)}$, the flattened feature is concatenated with the hallucination and the corresponding SSIM score, and then fed to a GRU. The output features are fed to a Gumbel Softmax \cite{Jang2017gumbel}, which makes the sampling vector differentiable.

Given the sampling vector, we now describe our frame-skipping routine. Denoting such number of skipping frames as $m^* = \argmax_m \br^{(t)}[m]$, there are two possible scenarios: $m^* = 0$ and $m^* \in [1, M]$. In the \textit{first} case, we do not skip anything and continue to run the remaining part of the network, thus the complexity is that of the full pipeline $\mathcal{O}_{full}$. In the \textit{second} case, we only pre-scan the current frame, which has already been done, and skip computation on the next $m^* - 1$ frames. The classification results and memory from recurrent models are propagated accordingly. The complexity for these $m^*$ frames is  $\mathcal{O}_{pre} = \mathcal{O}_{\phi_1^\lambda} + \mathcal{O}_{H} + \mathcal{O}_{\mathcal{T}}$, being the model's first half, hallucinator, and temporal sampler. Note that $\mathcal{O}_{full} = \mathcal{O}_{pre} + \mathcal{O}_{rest}$, where $\mathcal{O}_{rest}$ is the complexity of running the rest of the pipeline, including spatial sampler, other modalities, and classifier. Under such policy, we train the temporal sampler by minimizing the weighted sum of classification loss $\mathcal{L}_{class}$ (only using full-inference frames) and efficiency loss $\mathcal{L}_e$, 
which is defined as
\begin{equation}
    \mathcal{L}_e = n_{full} \cdot \mathcal{O}_{full} + n_{pre} \cdot \mathcal{O}_{pre},
    \label{equ:efficiency_loss}
\end{equation}
where $n_{full}$ and $n_{pre}$ are respectively the number of frames with full inference and only pre-scanning.
Without any constraints, it is possible that no frame would be fed to the second half of the pipeline, \ie, $\argmax_m \br^{(t)}[m] \neq 0, \forall t$. To avoid such scenario, we include a warm-up step, where the full pipeline is run at the first frame. It also helps initialize memory for recurrent models and ensures that we have classification result for at least one frame.

\section{Experiments}
\label{sec:experiment}

\begin{figure*}
    \centering
    \null\hfill
    \begin{subfigure}[t]{0.42\linewidth}
        \includegraphics[width=\linewidth]{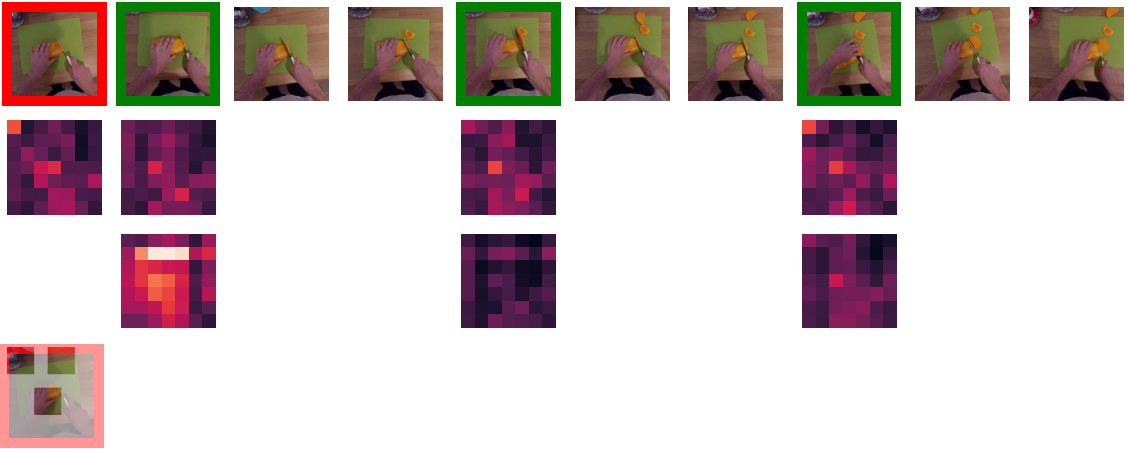}
        \caption{cut squash}
        \label{sfig:qualitative_a}
    \end{subfigure}
    \hfill
    \begin{subfigure}[t]{0.42\linewidth}
        \includegraphics[width=\linewidth]{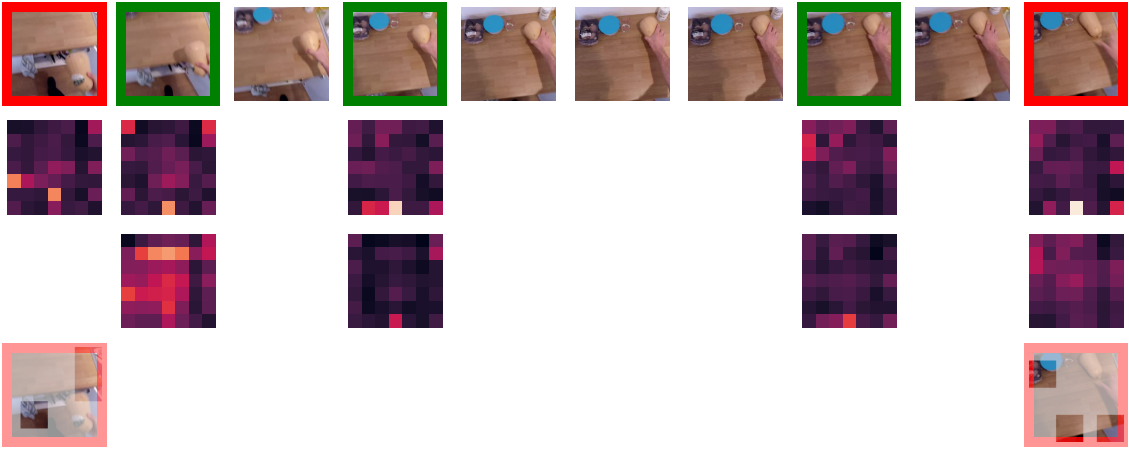}
        \caption{put squash}
        \label{sfig:qualitative_b}
    \end{subfigure}
    \hfill\null
    \\
    \null\hfill
    \begin{subfigure}[t]{0.42\linewidth}
        \includegraphics[width=\linewidth]{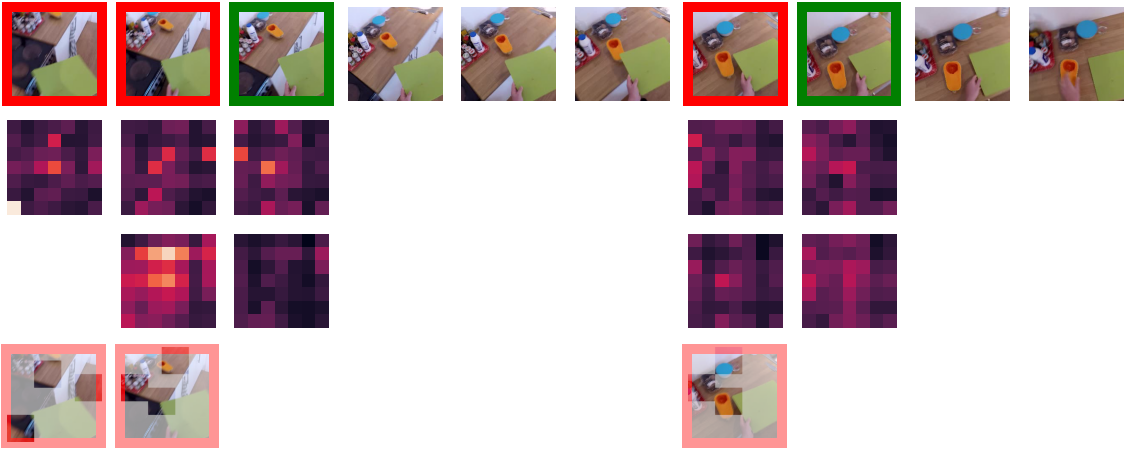}
        \caption{put board}
        \label{sfig:qualitative_c}
    \end{subfigure}
    \hfill
    \begin{subfigure}[t]{0.42\linewidth}
        \includegraphics[width=\linewidth]{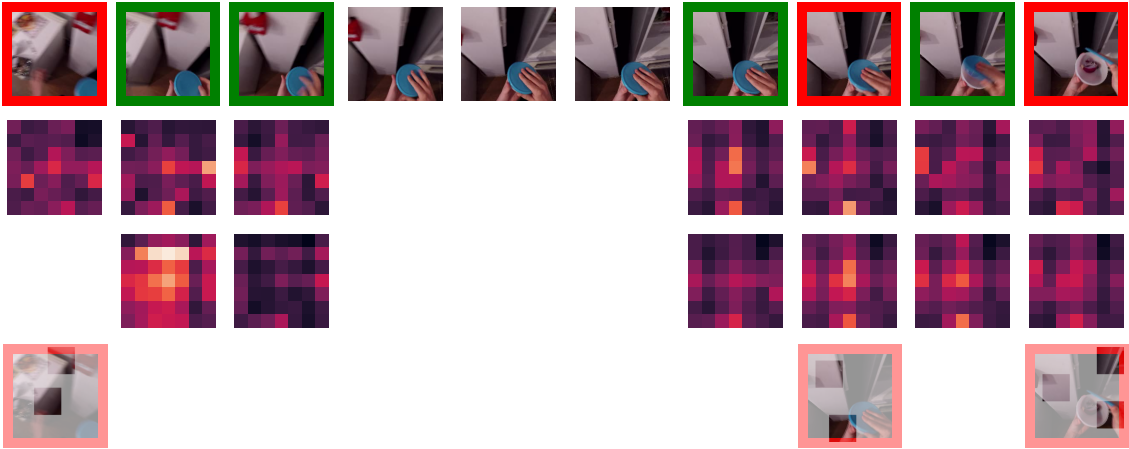}
        \caption{open container}
        \label{sfig:qualitative_d}
    \end{subfigure}
    \hfill\null
    \vspace{-.2cm}
    \caption{Qualitative results of spatiotemporal sampling on video sequences. From top to bottom are: the input video sequence, attention, hallucination generated from past attention, and the non-max suppressed top salient regions from the spatial sampler. The frames with red boundary are the non-skipped ones (full inference) while the frames with green boundary denote pre-scanning (running the first half). The frames without any boundary are the skipped ones. We choose to use the first frame to initiate the samplers and always associate it with the full inference.}
    \label{fig:qualitative}
    \vspace{-.5cm}
\end{figure*}

We evaluate our system on EPIC-KITCHENS 2018 \cite{damen2018epickitchens}, following the training and validation splits of \cite{kazakos2019tbn}, and split-1 of UCF-101 \cite{soomro2012ucf101}. EPIC-KITCHENS contains 55 hours of full-HD, 60fps egocentric videos, each of which is associated with a verb (125 classes) and a noun (331 classes) label. The action is thus defined as a pair of the corresponding verb and noun labels, \eg, [cut, squash] and [open, container]. UCF-101 is a dataset of generic actions, consisting of 27 hours of 25fps videos, divided into 101 action classes.

For EPIC-KITCHENS, we use two input modalities, namely \textit{RGB} and \textit{Spectrogram}, corresponding to the vision and audio domains. Although the optical flow is provided as a part of the dataset, we avoid using it because such data are computationally expensive in real-life scenarios. We treat RGB inputs as the guiding modality of the system because our hallucinator and samplers rely on the attention from vision data. The spectrogram inputs act as the additional modality and are only used when a frame is not skipped by the temporal sampler. For UCF-101, we only use the RGB modality for better comparison with our baseline.

We benchmark our system with top-1 and top-5 accuracy for both datasets. For EPIC-KITCHENS, we include the accuracy of three domains: action, verb, and noun. To assess the system's efficiency, we further report the models' FLOPS per frame, which is proportional to inference time and power consumption. Since the model complexity is time-variant for the experiments with temporal sampler, we instead provide the accumulated FLOPS over the whole validation sets and the average FLOPS per frame. We also report the trade-off factor, defined as GFLOPS per the top-1 accuracy, to compare efficiency of different models (lower means better). This metric indicate how much of computation is required for one percent of accuracy on average. 

\begin{table*}[]
    \centering
    \footnotesize
    {\setlength{\tabcolsep}{1.1em}
    \begin{tabular}{lcccccccc}
        \thickhline
        Model & Avg GFLOPS & Top-1 & Top-5 & Verb Top-1 & Verb Top-5 & Noun Top-1 & Noun Top-5 & Trade-off \\
        \hline \hline
        TBN \cite{kazakos2019tbn} (224) & 4.62 & 28.32 & 60.30 & 56.96 & 86.61 & 41.03 & 65.35 & 0.163 \\
        SAN19-baseline (224)            & 8.64 & 27.52 & 57.55 & 55.84 & 86.24 & 39.83 & 62.84 & 0.314 \\
        \hline
        $\mathcal{S}_0$ & \textbf{5.80} & 24.56 & 53.34 & 53.50 & 83.95 & 34.57 & 58.84 & \textbf{0.236} \\
        $\mathcal{S}_1$ & 6.16 & 25.23 & 54.05 & 55.17 & \textbf{84.49} & 35.49 & 59.68 & 0.244 \\
        $\mathcal{S}_2$ & 6.48 & 24.94 & 53.55 & 55.21 & 84.15 & 35.20 & 59.17 & 0.260 \\
        $\mathcal{S}_3$ & 6.80 & \textbf{25.77} & \textbf{54.42} & \textbf{55.71} & 84.15 & \textbf{35.78} & \textbf{59.84} & 0.264 \\
        \thickhline
    \end{tabular}}
    \vspace{-.2cm}
    \caption{Results of baselines and spatial sampler $\mathcal{S}$ on EPIC-KITCHENS validation set. The baseline models TBN \cite{kazakos2019tbn} and SAN19 takes 224x224 RGB inputs whereas our model uses 112x112 resolution RGB frames. The dimension of the Spectrogram is kept as 256x256. The average GFLOPS per-frame is included to indicate the model complexity. We showcase the performance as top-1 and top-5 per-video accuracy for action, verb, and noun domain. We also include the efficiency trade-off (lower means better), computed as the GFLOPS over top-1 accuracy, to show how much GFLOPS is needed with each accuracy percent on average. Our models with spatial samplers are denoted as $\mathcal{S}_k$, where $k$ is the number of RoIs extracted. $\mathcal{S}_0$ means no spatial sampling. $\mathcal{S}_3$ provides the best accuracy among spatial samplers and still with a lower complexity than the baseline.}
    \label{tab:spatial_sampler}
    \vspace{-.2cm}
\end{table*}

\begin{table*}[]
    \centering
    \footnotesize
    {\setlength{\tabcolsep}{1.05em}
    \begin{tabular}{lcccccccccccc}
        \thickhline
        Model & Prescan & Full & Total  & Avg    & Top-1 & Top-5 & Verb  & Verb  & Noun  & Noun  & Trade & Speed \\
              & (\%)    & (\%) & TFLOPS & GFLOPS &       &       & Top-1 & Top-5 & Top-1 & Top-5 & -off     & up    \\
        \hline \hline
        $\mathcal{S}_0$                &    0 & 100.00 & 139.14 & 5.80 & 24.56 & 53.34 & 53.50 & 83.95 & 34.57 & 58.84 & 0.236 & - \\
        \hline
        $\mathcal{S}_0, \mathcal{T}_1$ & 41.96 & 58.04 &  86.59 & 3.61 &         22.81  & 52.29 & 52.00 & 83.07 & 32.90 & \textbf{57.92} & 0.158 &         1.60x  \\
        $\mathcal{S}_1, \mathcal{T}_1$ & 49.17 & 50.83 &  \textbf{81.27} & \textbf{3.39} &         22.98  & 51.63 & 53.25 & 83.07 & 33.15 & 57.30 & \textbf{0.147} &         1.71x  \\
        $\mathcal{S}_2, \mathcal{T}_1$ & 49.96 & 50.04 &  83.96 & 3.50 &         23.52  & 52.09 & 53.34 & 82.32 & 32.90 & \textbf{57.92} & 0.149 &         1.76x  \\
        $\mathcal{S}_3, \mathcal{T}_1$ & \textbf{50.00} & \textbf{50.00} &  87.47 & 3.66 & \textbf{24.06} & \textbf{52.88} & \textbf{54.17} & \textbf{83.70} & \textbf{33.74} & \textbf{57.92} & 0.152 & \textbf{1.77x} \\
        \thickhline
    \end{tabular}}
    \vspace{-.2cm}
    \caption{Results of spatial sampler $\mathcal{S}$ and temporal samplers $\mathcal{T}$ on EPIC-KITCHENS validation set. The table includes the percentage of frames pre-scanned (\textit{Prescan (\%)}), and fully processed (\textit{Full (\%)}), the accumulated Tera-FLOPS over the whole validation set, and the average computation saving compared to its spatial sampler counterpart. All models have temporal sampling, except for the first row, which is copied from \tabref{tab:spatial_sampler} for comparison. All temporal samplers save the compute compared to $\mathcal{S}_0$ with tolerable loss of accuracy. The table only reports temporal sampling range of 1 ($\mathcal{T}_1$) so there are no skipped frames.}
    \label{tab:temporal_sampler}
    \vspace{-.4cm}
\end{table*}

\subsection{Implementation details}
\label{ssec:implementation_details}

\begin{table}[]
    \centering
    \footnotesize
    {\setlength{\tabcolsep}{0.85em}
    \begin{tabular}{lcccc}
        \thickhline
                        & $\mathcal{S}_0$ & $\mathcal{S}_1$ & $\mathcal{S}_2$ & $\mathcal{S}_3$ \\
        \hline \hline
        $\mathcal{T}_1$ &    \textbf{22.81}, 1.60x &    \textbf{22.98}, 1.71x &    \textbf{23.52}, 1.76x &    \textbf{\textit{24.06}}, 1.77x \\
        $\mathcal{T}_2$ &    22.52, 2.58x &    21.94, 2.44x &    22.23, 2.42x &    21.23, 2.62x \\
        $\mathcal{T}_3$ &    20.64, 3.29x &    21.06, 3.11x &    21.23, 3.21x &    20.73, 3.21x \\
        $\mathcal{T}_4$ &    18.85, \textbf{\textit{4.01x}} &    18.89, \textbf{3.64x} &    19.35, \textbf{3.43x} &    18.56, \textbf{3.92x} \\
        \thickhline
    \end{tabular}}
    \vspace{-.2cm}
    \caption{Accuracy and speed-up factors across different spatial samplers $\mathcal{S}$ and temporal samplers $\mathcal{T}$ on EPIC-KITCHENS validation set. Each column indicates number of regions $k$ for the spatial sampler $\mathcal{S}_k$ while each row describes the sampling range $M$ for temporal sampler $\mathcal{T}_M$. Each cell is a pair of top-1 accuracy and speed-up time corresponding to a spatiotemporal setting. Using spatial sampler improves the accuracy, but requires more complexity. Higher temporal sampling range $M$ corresponds to more speed-up, but also sacrifices more accuracy.}
    \label{tab:space_vs_time}
    \vspace{-.5cm}
\end{table}

Our system uses SAN19 with pairwise self-attention (equivalent to ResNet50 \cite{zhao2020san}) as the backbone network to extract features. For EPIC-KITCHENS, the additional Spectrogram (256x256) is constructed from the audio channels using the same processing procedure as in \cite{kazakos2019tbn}. We choose to extract attention at layer3-0 of the backbone network because it shows good trade-off between complexity and performance in our experiments. This gives us the attention feature map with the dimensionality of 32x7x7. Please refer to the supplement materials for additional analysis of picking layer 3-0 for the attention extraction.

The hallucinator is a Conv-LSTM with 1 layer and 32 hidden dimensions. It is equipped with a encoder and a decoder, each is a 2D Conv layer with kernel of size 3x3 and 32 channels. The action classifier used with our spatial and temporal sampler is a three-head GRU, corresponding to the global features (low-res RGB and Spectrogram), local features (cropped high-res RGB and Spectrogram), and their concatenation to the master GRU head. The goal of the multi-head architecture is to ensure the network extract prominent features from the cropped regions rather than relying solely on the low-res image. Each head of the GRU classifier and our GRU temporal sampler share the same architecture of 2 layers and 1024 hidden dimension.

The whole system is trained in multiple phases. We first train the two feature extraction modules with FC classifier, corresponding to the low-res and high-res inputs. We train the models with the standard cross-entropy loss for 100 epochs, using SGD with momentum of 0.9 \cite{qian1999momentum}, with decaying at epochs 30, 60, and 90. The weights of feature extraction modules are frozen and used for other models. The hallucinator is then trained using the belief loss $\mathcal{L}_b$ in \equref{equ:belief_loss} with teacher forcing routine \cite{williams1989teacherforcing}. For the spatial sampler with three-head classifier, the predictions on all heads are averaged and the model is trained using the loss $\mathcal{L}_{class} = \sum_{h=1}^3 \theta_h \mathcal{L}_h$, where $\mathcal{L}_h$ is the cross-entropy loss of a head and $\theta_h$ is the corresponding scaling. The temporal sampler is jointly trained with the pretrained three-head classifier and the fixed spatial sampler, using the total loss $\mathcal{L}_{class} + \theta_e\mathcal{L}_e$, where $\mathcal{L}_e$ is the efficiency loss described in \equref{equ:efficiency_loss} with the corresponding scaling $\theta_e$. We train each sampling model for 50 epochs using Adam optimizer \cite{kingma2015adam}. For feature extraction modules, we only sample three frames since we only aim to extract spatial features instead of temporal ones in this phase. As for the sampling modules, we use 10 frames for EPIC-KITCHENS and 16 frames for UCF-101 as they requires more temporal information.

\subsection{Qualitative results}
\label{ssec:qualitative_results}

We demonstrate our qualitative results from EPIC-KITCHENS dataset in \figref{fig:qualitative} to show the outputs of both spatial and temporal samplers. The frames are uniformly sampled from the validation videos. We use ``red'' and ``green' color to highlight full inference or simply pre-scanning. Unmarked frames are skipped without any computation. The spatial sampler only runs on ``red'' frames to enrich data, therefore the cropped regions are only available for these frames. We choose to use $\mathcal{S}_3$ and $\mathcal{T}_4$ in qualitative experiments since they provide more observable sampling results for visualization, regardless of their effect on the qualitative performance.

Overall, the temporal sampler can adaptively sample the frames. The number of ``red'' frames is fewer than the original video length and can compactly describe the complete action. In \figref{sfig:qualitative_a}, the action cutting squash is a simple example since it can be easily represented using a single frame. We see that aside from the warming up first frame, the temporal sampler here only pre-scans three frames and skip the rest of them. The action of putting down a squash in \figref{sfig:qualitative_b} is another interesting example, where the first and last frame are selected, corresponding to when the actor is holding the squash in hand and placing it on the chopping board. These two sampled frames concisely represent the action putting down is reality. A similar example is illustrated in \figref{sfig:qualitative_c}, where the actor is putting down a chopping board. \figref{sfig:qualitative_d} depicts a more challenging video sequence of opening a container, as the background is not informative and the container are not opened until the final frame, resulting in more consecutive pre-scanned frames.

\subsection{Quantitative results}
\label{ssec:quantitative_results}

\begin{table}[]
    \centering
    \footnotesize
    {\setlength{\tabcolsep}{1em}
    \begin{tabular}{lcccc}
        \thickhline
        Model & Avg    & Top-1 & Top-5 & Trade \\
              & GFLOPS &       &       & -off  \\
        \hline \hline
        TSN \cite{wang2016tsn}(224) & 4.12 & 80.94 & 95.66 & 0.051 \\
        SAN19-baseline(224)                  & 3.75 & 80.57 & 94.08 & 0.047 \\
        \hline
        $\mathcal{S}_0$              & \textbf{0.90} & 69.81 & 90.11 & \textbf{0.013} \\ 
        $\mathcal{S}_1$              & 1.23 & 72.14 & 90.59 & 0.017 \\
        $\mathcal{S}_2$              & 1.55 & \textbf{72.19} & 90.62 & 0.021 \\
        $\mathcal{S}_3$              & 1.87 & 72.03 & \textbf{91.15} & 0.026 \\
        \thickhline
    \end{tabular}}
    \vspace{-.2cm}
    \caption{Results of baselines and spatial samplers $\mathcal{S}$ on UCF-101 (split-1). In the first two rows, we reproduce TSN \cite{wang2016tsn} results for split-1 of UCF-101 and compare with our backbone SAN19 using 224x224 RGB inputs. The other models uses 112x112 frames. We evaluate our systems using top-1 and top-5 accuracy, together with average GFLOPS per-frame and efficiency trade-off factors, similar to \tabref{tab:spatial_sampler}. Overall, we obtain the best $\mathcal{S}_2$ among all spatial samplers with significantly less GFLOPS than the baselines.}
    \label{tab:ucf_101_space}
    \vspace{-.1cm}
\end{table}

\begin{table}[]
    \centering
    \footnotesize
    {\setlength{\tabcolsep}{0.7em}
    \begin{tabular}{lcccccc}
        \thickhline
        {\footnotesize Model} & {\footnotesize Total}  & {\footnotesize Avg}    & {\footnotesize Top-1} & {\footnotesize Top-5} & {\footnotesize Trade} & {\footnotesize Speed} \\
              & {\footnotesize TFLOPS} & {\footnotesize GFLOPS} &       &       & {\footnotesize -off}  & {\footnotesize -up}   \\
        \hline \hline
        $\mathcal{S}_0, \mathcal{T}_1$ & \textbf{44.56} & \textbf{0.74} & 70.21 & 89.96 & \textbf{0.010} & 1.23x \\
        $\mathcal{S}_1, \mathcal{T}_1$ & 54.61 & 0.90 & \textbf{71.43} & 90.48 & 0.013 & 1.37x \\
        $\mathcal{S}_2, \mathcal{T}_1$ & 64.48 & 1.07 & 71.19 & 90.72 & 0.015 & 1.46x \\
        $\mathcal{S}_3, \mathcal{T}_1$ & 74.50 & 1.23 & 71.21 & \textbf{90.75} & 0.017 & \textbf{1.52x} \\
        \thickhline
    \end{tabular}}
    \vspace{-.2cm}
    \caption{Results of baselines and spatial samplers and temporal samplers on UCF-101 (split-1). We achieve the best results by combining $\mathcal{S}_3$ with $\mathcal{T}_1$. Our sampling routine provides good speed up compared to their spatial-sampler-only counterparts, while still retaining comparable accuracy.}
    \label{tab:ucf_101_time}
    \vspace{-.4cm}
\end{table}

\tabref{tab:spatial_sampler} shows the quantitative results of our spatial sampler on EPIC-KITCHENS. The first two rows are TBN \cite{kazakos2019tbn} and SAN19-baseline with FC classifier, both use high-res RGB (224x224) and Spectrogram (256x256). Since TBN relies on Inception backbone, its model complexity is not directly comparable with our experiments, with use SAN19 backbone. However, our baseline model provides comparable accuracy. Since the main objective of the paper is to increase efficiency of a given model, we focus on comparing performance and complexity with the baseline SAN19.

The rest of \tabref{tab:spatial_sampler} shows our results of the spatial sampler with GRU classifier, using the low-res RGB (112x112), cropped high-res RGB (64x64), and the same Spectrogram inputs. We denote $\mathcal{S}_k$ as our spatial sampler with top-$k$ RoIs, where $\mathcal{S}_0$ means no spatial sampling involved. It can be seen that by simply decreasing the image resolution,  $\mathcal{S}_0$ can reduce the complexity by 2.84 GFLOPS with a small loss of 2.96\% in accuracy. By adding the sampled regions from the spatial sampler $\mathcal{S}_1$, $\mathcal{S}_2$, and $\mathcal{S}_3$, the accuracy is consistently improved. We achieve the best performance with $\mathcal{S}_3$ across our spatial sampling experiments. This gets close to the performance of the baseline, with 1.75\% accuracy different but still can save 1.84 GFLOPS of computation. Furthermore, the efficiency trade-off of $\mathcal{S}_3$ is lower, showing that the model with spatial sampler is more efficient in terms of average GFLOPS per accuracy.

\tabref{tab:temporal_sampler} shows our results with both spatial and temporal samplers $\mathcal{T}_1$. For convenient comparison, we include the results of $\mathcal{S}_0$ from \tabref{tab:spatial_sampler} with its accumulated TFLOPS over the whole validation set. We observe no skipping frames in $\mathcal{T}_1$ because this model only allows either pre-scanning or full-inference. \tabref{tab:space_vs_time} shows the performance across different choices of spatial and temporal samplers. Each pair of items in the table represent top-1 accuracy and speed-up time corresponding to a set of spatiotemporal sampling parameters. Compared to the spatial-sampler-only counterparts, $\mathcal{T}_4$ can reduce the complexity up to 4.01 times by sacrificing more accuracy. On the other end of the spectrum, $\mathcal{T}_1$ can approximate the original accuracy, and still with speed-up time up to 1.77x. For more detailed results and analysis between spatial and temporal samplers on EPIC-KITCHENS, please refer to our supplementary materials.

We report the results of UCF-101 in \tabref{tab:ucf_101_space} and \tabref{tab:ucf_101_time}, following similar convention as in \tabref{tab:spatial_sampler} and \tabref{tab:temporal_sampler}. We reproduce the results of TSN \cite{wang2016tsn} on split-1 of the dataset using our hardware for more comparable results. Behaviors similar to EPIC-KITCHENS are observed in this dataset, \ie, the both spatial and temporal samplers can reduce complexity while maintaining comparable accuracy. $\mathcal{S}_3, \mathcal{T}_1$ has the highest speed-up with only 0.82 loss of top-1 accuracy and $\mathcal{S}_1,\mathcal{T}_1$ has the best accuracy with 1.37x speed-up.

\section{Conclusion}
\label{sec:conclusion}

We introduce an attention-based spatiotemporal sampling scheme to adaptively sample videos for efficient action recognition. Spatial sampler provides a global view at low-res and local salient views at high-res. Temporal sampler pre-scans and decides the sampling strategy by comparing the current attention with the past hallucination. Future work includes exploring more modalities to improve system performance and flexibly choosing layers of attention.

{\small
\bibliographystyle{ieee_fullname}
\bibliography{main.bib}
}

\newpage
\onecolumn

\appendix
\begin{center}
    \Large
    \textbf{Supplementary Materials for Efficient Human Vision Inspired Action Recognition using Adaptive Spatiotemporal Sampling}
\end{center}
\addcontentsline{toc}{section}{Appendices}
\renewcommand{\thesubsection}{\Alph{subsection}}

\section{Foveal vision in human visual perception and spatial sampling}

\begin{figure}[h]
	\centering
	\includegraphics[width=\linewidth]{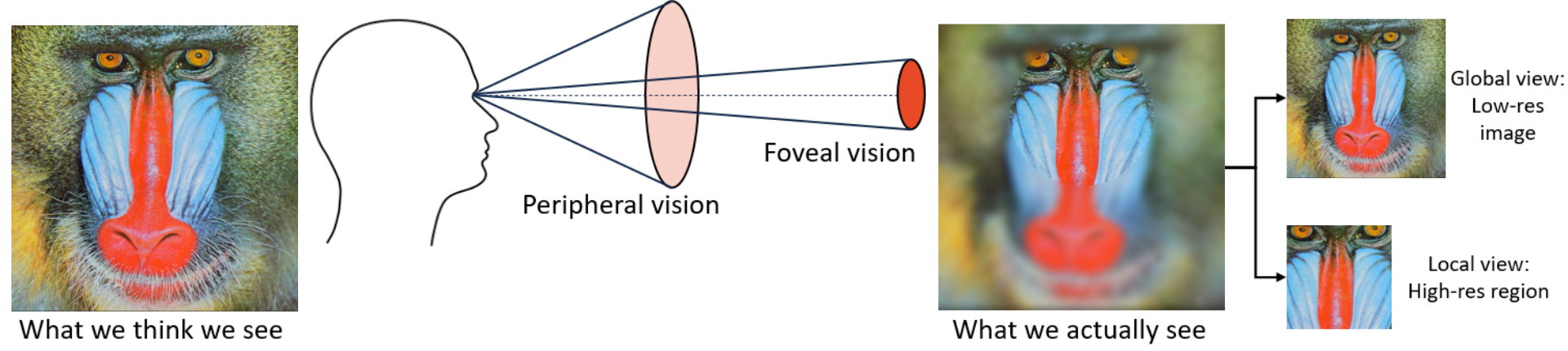}
	\caption{The foveal vision in human corresponds to sharp and centered vision to obtain fine local details, while the peripheral vision corresponds to low visual acuity to get coarser global information. This is similar to the local-view from high-res region and global-view from low-res image in our spatial sampler.}
	\label{fig:foveal_vision}
\end{figure}

\figref{fig:foveal_vision} explains the concepts of foveal vision. Although our eyes are often compared to a camera, processing across the visual field is quite different. Our visual field consists of at least two parts, namely the foveal vision gives us the fine local details, and the peripheral vision gives us a coarser global view. Therefore, given the example picture of a baboon, the left image is not actually what we see. Instead, what we actually see is closer to the image on the right, sharp at the center and blurry at the boundary.

Going back to computers, we can simulate the pair of foveal vision and peripheral vision as two views of an image, where the global view correspond to low resolution input, that capture the overall detail. The local view corresponds to high resolution input, that captures the important details. Here, we use the local view with smaller size but with the same resolution.

While we can direct our gaze to important regions so that they are always located in our central vision, it is not possible to do the same thing for a given captured image. To determine the region, the proposed system relies on the salient regions from attention maps instead, which guide the spatial sampling routine.

\section{Pre-attentive processing in human visual perception and temporal sampling}

\begin{figure}[h]
	\centering
	\includegraphics[width=0.9\linewidth]{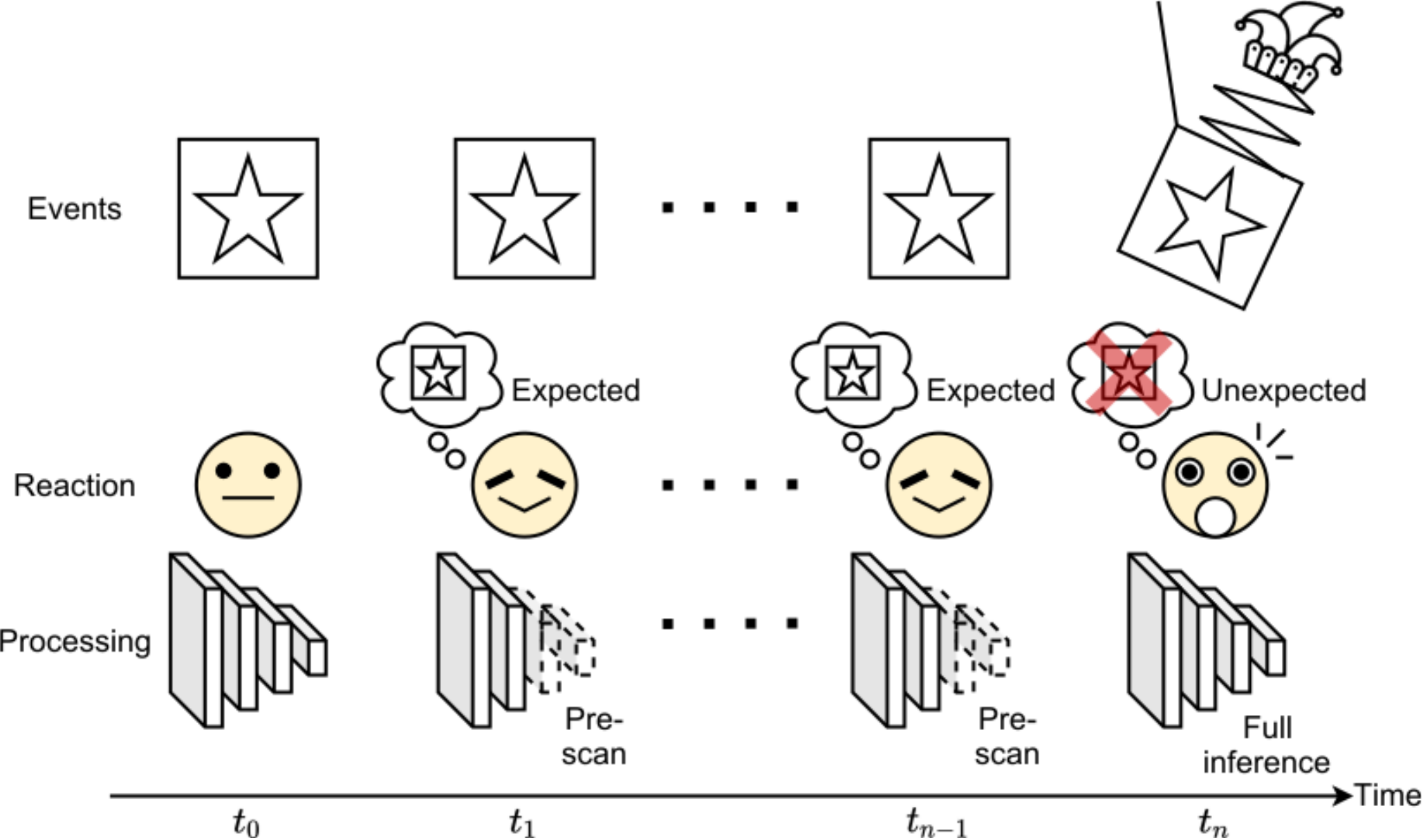}
	\caption{Pre-attentive processing across time domain. Getting bored is a by-product of observing expected events, while getting surprised is the result of seeing something unexpected. This is similar to pre-scan new frames to see if they are similar to the hallucinated prediction in our temporal sampler.}
	\label{fig:preattentive_processing}
\end{figure}

Pre-attentive processing is a subconscious accumulation of info from the environment, \ie, all available information is pre-attentively processed, then our brains will choose the important event to dive deep in. Motion is also a pre-attentive feature, therefore this mechanism inspires our temporal sampling scheme.

A by-product of pre-attentive processing is that predictable events are more likely to be ignored. In \figref{fig:preattentive_processing}, we look at an example of a box over time. While observing the same thing continuously, humans tend to get bored since our brains can predict that nothing unexpected is happening. In such case, we do not have to use a much processing power. However, if something unexpected happens, \eg, the Jack-in-the-box appears in the last frame of the example, we will be surprised and immediately activate our sensors to process the new event, since it is different from our expectation.

Furthermore, in deep networks, we know that the first few layers can already capture some kind of features. Therefore, we can use those layers to pre-scan new frames to determine the expectation. Specifically, if they give similar features as the past, we do not need to process any further, meaning there is no need for the full model inference. Such expectation of the future is modeled by our hallucinator, where the results are used to determine the sampling strategy in the temporal sampler.

\section{Choosing layer to extract attention}

\begin{figure}[h]
	\centering
	\includegraphics[width=0.8\linewidth]{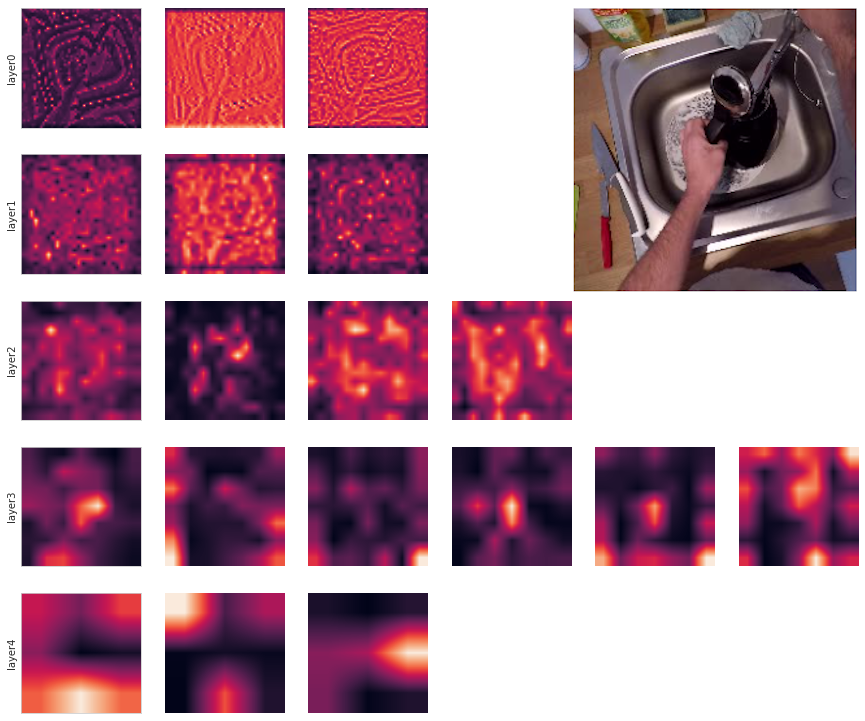}
	\caption{Attention extracted from all bottleneck layers of SAN19. The input image is showed on the top-right corner. Earlier layers show more fragmented regions while latter ones provide more concise attentions. We visualize the attention maps with bilinear interpolation for better visibility among different layers. The color mapping of the visualization is not normalized because of different range of values across layers.}
	\label{fig:attention_all_layers}
\end{figure}

We explain why we choose to use layer3-0 to extract attention in this section. \figref{fig:attention_all_layers} shows the attentions from all bottleneck layers of SAN19, where the input with size of 3x112x112 is provided on the top-right corner of the figure. Since SAN19 is equivalent to ResNet50, there are five different blocks, named as layer0 to layer4. Each layer block here has different modules, namely layer$x$-$y$, where $x$ and $y$ are the layer and module indices, respectively. The dimensionality of all attention maps in \figref{fig:attention_all_layers} is as follow:
\begin{itemize}
    \item Layer0-0, layer0-1, layer0-2: 2x56x56
    \item Layer1-0, layer1-1, layer1-2: 8x28x28
    \item Layer2-0, layer2-1, layer2-2, layer2-3: 16x14x14
    \item Layer3-0, layer3-1, layer3-2, layer3-3, layer3-4, layer3-5: 32x7x7
    \item Layer4-0, layer4-1, layer4-2: 64x3x3
\end{itemize}

\begin{figure}[h]
	\centering
	\includegraphics[width=0.9\linewidth]{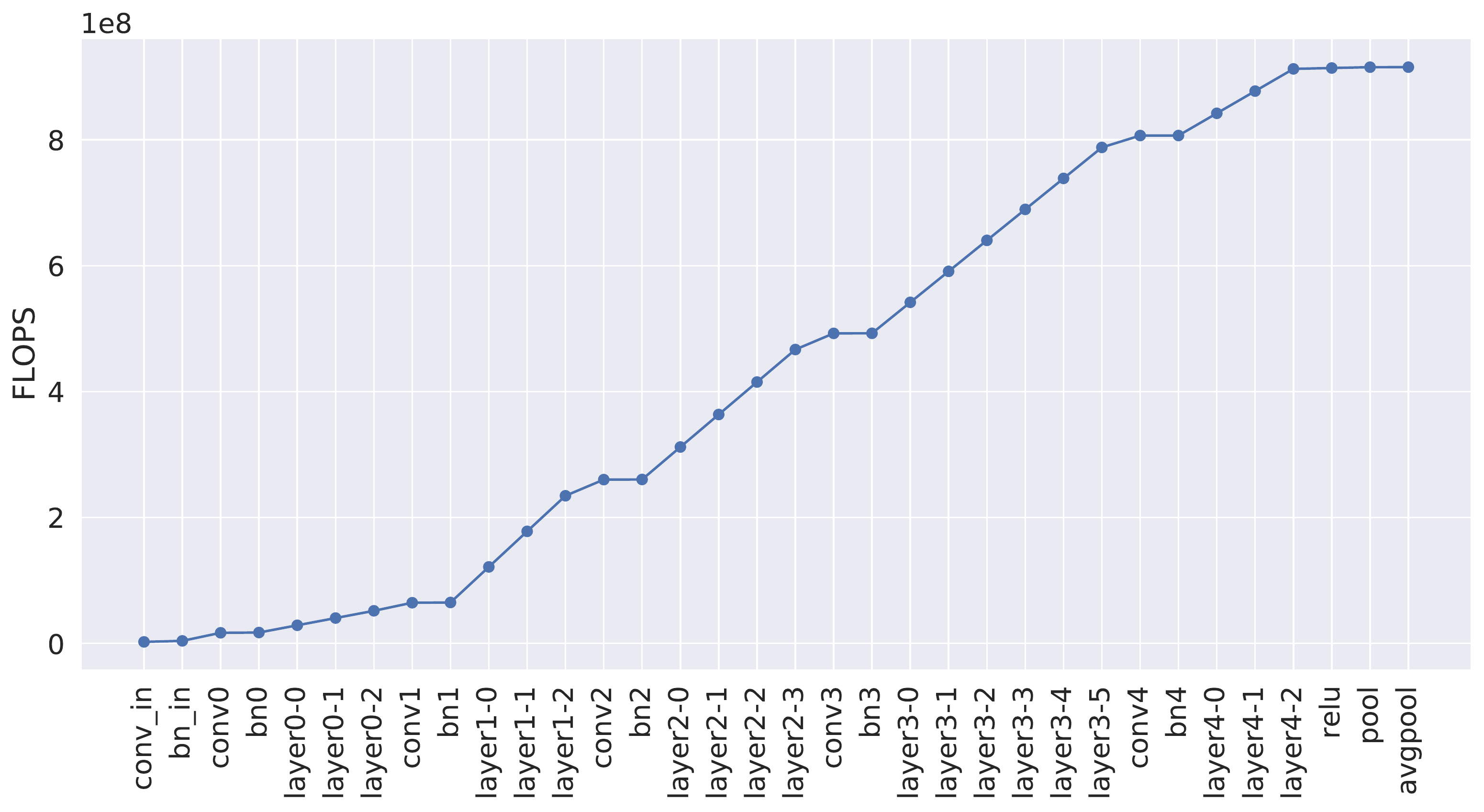}
	\caption{Accumulated complexity up to different layers of SAN19. The input size is fixed as 3x112x112.}
	\label{fig:layer_flops_analysis}
\end{figure}

The more layers we use, the more information is encoded in the corresponding features. Thus the associated attention maps also carry more information. We see that the attention maps in layer0, layer1, and layer2 are noisier, while the layer3 and layer4 give cleaner and more interpretable attention. For the temporal sampler, attention map with complicated pattern will make it more difficult to hallucinate the future attention. On the other hand, using more layers results in smaller spatial dimension and more complexity. A tiny attention can make it challenging for the spatial sampler to select the correct region of interest. As for model complexity, \figref{fig:layer_flops_analysis} shows our analysis on the same model and input dimension. The horizontal axis shows the layer names of the model while the vertical axis indicates the accumulated FLOPS up to that layer. We see that inference up to layer3-0 takes around half of the whole model complexity. Therefore, we decide to use attention at layer3-0 as it has good trade-off between complexity and quality for our system.

\section{Cumulative global attention and the sliding effect in neighboring footprints}

\begin{figure}[h]
	\centering
	\includegraphics[width=0.6\linewidth]{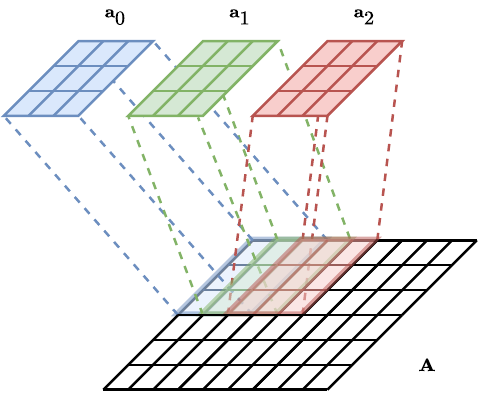}
	\caption{Cumulative global attention addresses the sliding effects. The top row is local attention associated with some neighboring footprints. The bottom row shows the global attention aggregated from the local attentions. The sliding effect in this example is similar to how to window kernel moves in convolution (with stride of 1).}
	\label{fig:sliding}
\end{figure}

\begin{figure}[h]
    \centering
    \begin{subfigure}[h]{0.3\linewidth}
        \includegraphics[width=\linewidth]{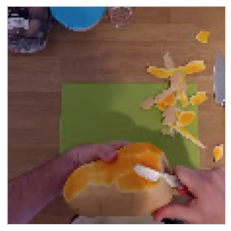}
        \caption{Input image.}
        \label{sfig:global_attention_input}
        \includegraphics[width=\linewidth]{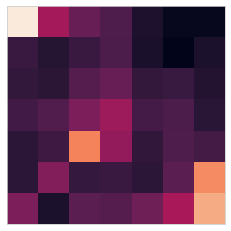}
        \caption{Cumulative global attention $\mathbf{A}$.}
        \label{sfig:global_attention_global}
    \end{subfigure}
    \begin{subfigure}[h]{0.675\linewidth}
        \includegraphics[width=\linewidth]{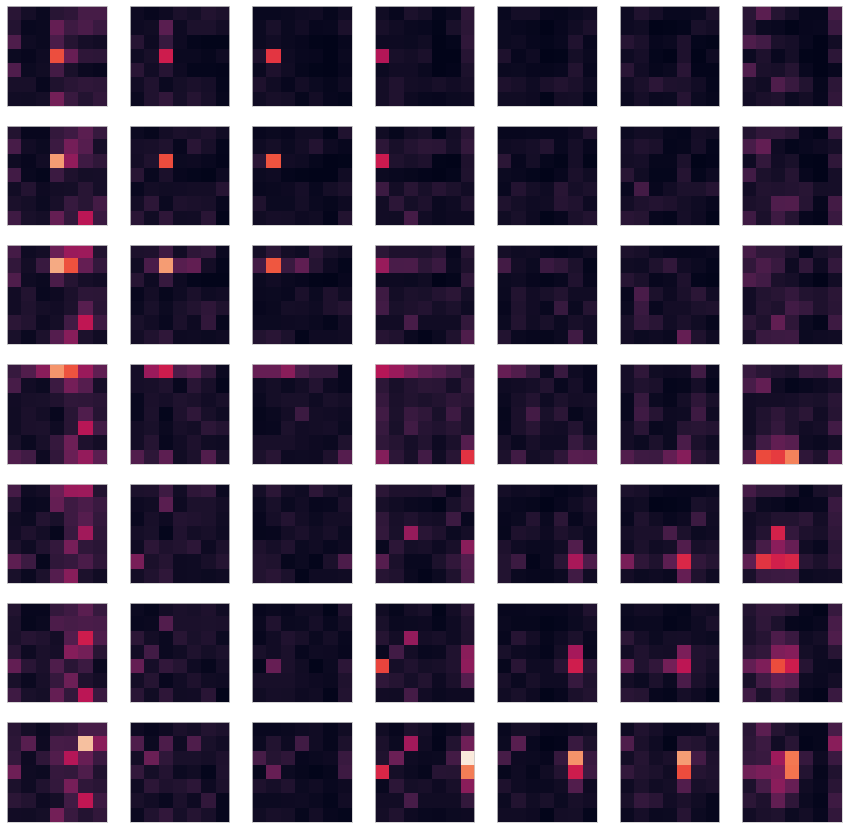}
        \caption{Local attentions $\ba_i$ from different footprint $\mathcal{R}_i$.}
        \label{sfig:global_attention_local}
    \end{subfigure}
    \caption{Local attention and cumulative global attention from the same input image and at the same layer. Hotter color indicates more salient regions. We average across the channel dimension for visualization.}
    \label{fig:global_attention}
\end{figure}

This section explains the cumulative global attention with further detail. We know that each local attention $\ba_i$ is defined with respect to the footprint $\mathcal{R}_i$, as shown in \equref{equ:local_attention}. The locations of neighboring footprints here are similar result in overlapping regions similar to the concept of moving the kernel window in convolution. Such overlapping causes the sliding effect, which is observed in neighboring $\ba_i$ in \figref{fig:global_attention}. Therefore, we can construct a global view of the attention by aggregating the local attention similarly to convolution to remove such sliding effect, as seen in \figref{fig:sliding}. This method creates duplication over the overlapping regions, \ie,
\begin{equation}
    \bA^* = \bA \odot M,
\end{equation}
where $\bA$ is the cumulative global attention defined in \equref{equ:global_attention}, $\bA^*$ is the clean global attention without overlapping, and $M$ is the mask counting such duplication. However, we see that the counter mask $M$ is a constant defined by the size of $\bA$ and $\ba$. Therefore, using $\bA$ instead of $\bA^*$ does not affect the quality of our hallucinator.

\figref{fig:global_attention} shows an example of the cumulative global attention (\figref{sfig:global_attention_global}), aggregated from the local attentions across multiple footprints (\figref{sfig:global_attention_local}), given the same input (\figref{sfig:global_attention_input}). We see that the neighboring $\ba_i$ have some sliding effect because of how we move the footprints, similar to convolution. Such effect is already encoded in $\bA$, making it easier to learn. Specifically, the activation of $\bA$ reflects the ``important regions'' in the input images, being the hands and the bowl (top-left corner). It motivates to use such global attention maps to find ROIs in our spatial sampler.

\section{Connected regions in spatial sampler}

\begin{figure}[h]
	\centering
	\includegraphics[width=\linewidth]{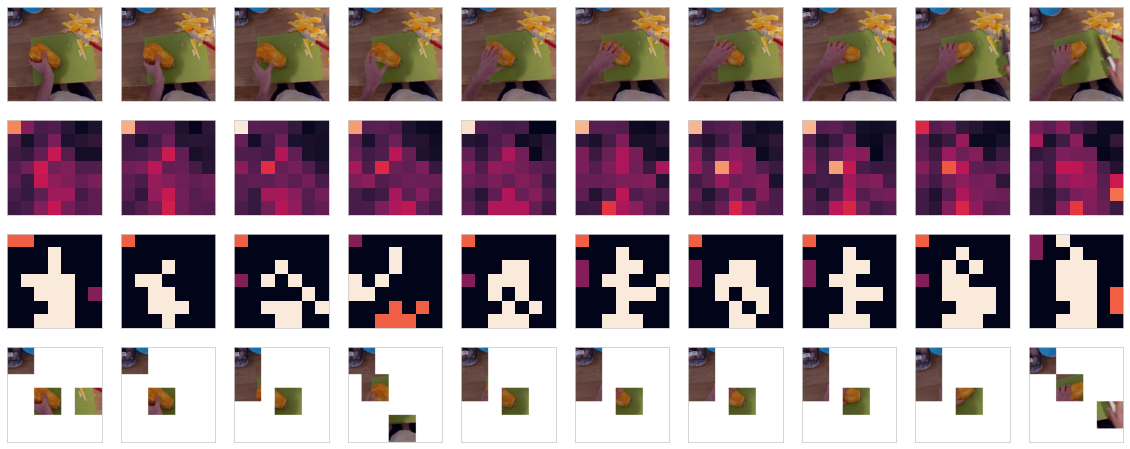}
	\caption{Detail of the spatial sampler. The first row is the input RGB frames and the second row is the corresponding attention maps. The third row shows the masks of the top-3 regions, where each color denotes a different region. The bottom row illustrates the sampled regions, corresponding to centroids of each mask.}
	\label{fig:spatial_mask}
\end{figure}

\figref{fig:spatial_mask} shows the details of our spatial sampler. Giving an attention from a input image, the sampler suppresses the low values to retrieve the regions with high saliency. However, this could result in multiple fragmented regions. To avoid this, we find the connected regions with 2-connectivity, \ie, non-suppressed pixels $[x_1, y_1]$ and $[x_2, y_2]$ are considered to belong to the same region if $|x1 - x2| + |y1 - y2| \leq 2$. We then select the top-$k$ regions based on the its score, defined as sum of all pixels inside the region. Such score scales according to both the values and the size of a region. The results of this procedure is illustrated in the third row of \figref{fig:spatial_mask}. The centroids of such regions (in attention plane) are projected by scaling to find the corresponding centroids in image plane. The sampler then find the bounding boxes surrounding such centroids to generate the regions of interest, as shown in the bottom row of \figref{fig:spatial_mask}.

\section{Spatial sampler results with smaller input size}

We provide additional results from the spatial sampler with smaller input size of 64x64 instead of 112x112 as in the main part of the paper. It means that both of the global and local views now share the same input dimension. The objective of this experiment is to see how much computation we can save by extensively reducing input size and how much accuracy is compromised. \tabref{tab:spatial_sampler_64} shows the results of this experiments, with the same convention as in \tabref{tab:spatial_sampler}. We see that using lower input dimension helps consistently reduce the model complexity by 0.59 GFLOPS, but the top-1 accuracy is reduced by 2.89\% on average. The trade-off factors here are also higher than the 112x112 experiments. Therefore, using input with too low dimension hurts the overall performance instead.

\begin{table}[h]
    \centering
    \small
    \begin{tabularx}{\textwidth}{Xcccccccc}
        \thickhline
        Model & Avg GFLOPS & Top-1 & Top-5 & Verb Top-1 & Verb Top-5 & Noun Top-1 & Noun Top-5 & Trade-off \\
        \hline \hline
        $\mathcal{S}_0$     & \textbf{5.22} & 21.77 & 49.04 & 51.54 & 83.11 & 30.61 & 54.92 & \textbf{0.240} \\
        $\mathcal{S}_1$     & 5.57 & 22.19 & 48.62 & 52.92 & 82.49 & 30.48 & 53.63 & 0.251 \\
        $\mathcal{S}_2$     & 5.89 & \textbf{22.64} & \textbf{50.08} & 52.96 & \textbf{83.32} & 31.78 & \textbf{55.25} & 0.260 \\
        $\mathcal{S}_3$     & 6.21 & 22.35 & 49.67 & \textbf{53.21} & \textbf{83.32} & \textbf{32.07} & \textbf{55.25} & 0.278 \\
        \thickhline
    \end{tabularx}
    \caption{Results of spatial sampler $\mathcal{S}$ on validation set of EPIC-KITCHENS with smaller input size of 64x64 for the global view. Overall, the model complexity is reduced with compromised performance. However, the trad-off factors are higher than using input size of 112x112.}
    \label{tab:spatial_sampler_64}
\end{table}

\section{Temporal sampler results with other sampling ranges}

\begin{table}[h]
	\centering
	\small
	\begin{tabularx}{\textwidth}{Xccccccccccccc}
		\thickhline
		Model                          &  Skip & Scan  & Full   & Total  & Avg    &         Top-1 & Top-5 & Verb  & Verb  & Noun  & Noun  & Trade & Speed          \\
									   &  (\%) & (\%)  & (\%)   & TFLOPS & GFLOPS &               &       & Top-1 & Top-5 & Top-1 & Top-5 & -off  & up             \\
		\hline \hline
		$\mathcal{S}_0$                &  0.00 &  0.00 & 100.00 & 139.14 &   5.80 &        24.56  & 53.34 & 53.50 & 83.95 & 34.57 & 58.84 & 0.236 & -              \\
		\hline
		$\mathcal{S}_0, \mathcal{T}_1$ &  0.00 & 41.96 & 58.04 &  86.59 &   3.61 &         22.81  & 52.29 & 52.00 & 83.07 & 32.90 & 57.92 & 0.158 &         1.60x  \\
		$\mathcal{S}_1, \mathcal{T}_1$ &  0.00 & 49.17 & 50.83 &  81.27 &   3.39 &         22.98  & 51.63 & 53.25 & 83.07 & 33.15 & 57.30 & 0.147 &         1.71x  \\
		$\mathcal{S}_2, \mathcal{T}_1$ &  0.00 & 49.96 & 50.04 &  83.96 &   3.50 &         23.52  & 52.09 & 53.34 & 82.32 & 32.90 & 57.92 & 0.149 &         1.76x  \\
		$\mathcal{S}_3, \mathcal{T}_1$ &  0.00 & 50.00 & 50.00 &  87.47 &   3.66 & \textbf{24.06} & 52.88 & 54.17 & 83.70 & 33.74 & 57.92 & 0.152 & \textbf{1.77x} \\
		\hline
		$\mathcal{S}_0, \mathcal{T}_2$ & 14.05 & 52.34 & 33.61 &  53.82 &   2.24 & \textbf{22.52} & 50.75 & 51.59 & 82.61 & 32.15 & 56.84 & 0.100 &         2.58x  \\
		$\mathcal{S}_1, \mathcal{T}_2$ & 14.35 & 51.58 & 34.07 &  56.91 &   2.37 &         21.94  & 51.50 & 51.50 & 83.32 & 33.24 & 56.88 & 0.108 &         2.44x  \\
		$\mathcal{S}_2, \mathcal{T}_2$ & 12.74 & 52.18 & 35.08 &  61.11 &   2.55 &         22.23  & 51.92 & 51.92 & 83.28 & 32.03 & 57.59 & 0.115 &         2.42x  \\
		$\mathcal{S}_3, \mathcal{T}_2$ & 15.02 & 52.70 & 32.28 &  59.27 &   2.47 &         21.23  & 51.50 & 48.92 & 83.28 & 32.36 & 56.88 & 0.116 & \textbf{2.62x} \\
		\hline
		$\mathcal{S}_0, \mathcal{T}_3$ & 25.99 & 48.36 & 25.65 &  42.19 &   1.76 &         20.64  & 49.37 & 49.21 & 82.40 & 30.28 & 55.00 & 0.085 & \textbf{3.29x} \\
		$\mathcal{S}_1, \mathcal{T}_3$ & 25.46 & 48.54 & 25.99 &  44.63 &   1.86 &         21.06  & 50.29 & 50.00 & 82.99 & 31.86 & 56.09 & 0.088 &         3.11x  \\
		$\mathcal{S}_2, \mathcal{T}_3$ & 25.75 & 48.60 & 25.64 &  46.04 &   1.92 & \textbf{21.23} & 51.71 & 49.57 & 83.28 & 31.23 & 57.51 & 0.090 &         3.21x  \\
		$\mathcal{S}_3, \mathcal{T}_3$ & 25.18 & 48.92 & 25.89 &  48.41 &   2.02 &         20.73  & 50.67 & 49.99 & 83.20 & 31.19 & 56.24 & 0.097 &         3.21x  \\
		\hline
		$\mathcal{S}_0, \mathcal{T}_4$ & 34.80 & 44.65 & 20.55 &  34.58 &   1.44 &         18.85  & 48.83 & 47.16 & 81.78 & 29.11 & 54.67 & 0.077 & \textbf{4.01x} \\
		$\mathcal{S}_1, \mathcal{T}_4$ & 32.06 & 46.16 & 21.78 &  38.12 &   1.59 &         18.89  & 49.12 & 48.79 & 81.90 & 30.15 & 55.34 & 0.084 &         3.64x  \\
		$\mathcal{S}_2, \mathcal{T}_4$ & 35.53 & 40.03 & 24.44 &  43.05 &   1.80 & \textbf{19.35} & 49.42 & 48.29 & 82.03 & 30.61 & 55.76 & 0.093 &         3.43x  \\
		$\mathcal{S}_3, \mathcal{T}_4$ & 34.63 & 44.52 & 20.85 &  39.66 &   1.65 &         18.56  & 47.50 & 47.00 & 81.61 & 29.36 & 54.34 & 0.089 &         3.92x  \\
		\thickhline
	\end{tabularx}
	\caption{Results of spatial sampler $\mathcal{S}$ and temporal samplers $\mathcal{T}$ on EPIC-KITCHENS validation set. Each block show the results corresponding to the temporal sampler $\mathcal{T}_M$, where $M$ is the maximum number of frames that $\mathcal{T}$ allows to skip. We highlight the best top-1 accuracy and speeding up factor for each $\mathcal{T}_M$ block. The table includes the percentage of frames being skipped (\textit{skip \%}), pre-scanned (\textit{scan \%}), and not skipped (\textit{full \%}), the accumulated Tera-FLOPS over the whole validation set, and the average computation saving compared to its spatial sampler counterpart. All models have temporal sampling, except for the first row, which is copied from \tabref{tab:spatial_sampler} for comparison. All temporal samplers significantly save the compute compared to the baseline system.}
	\label{tab:temporal_sampler_full}
\end{table}

\tabref{tab:temporal_sampler_full} shows our results with both spatial and temporal samplers. For better comparison, we include the results of $\mathcal{S}_0$ from \tabref{tab:spatial_sampler} with its accumulated TFLOPS over the whole validation set. Notice that the number of skipped and pre-scanned frames are both zero for $\mathcal{S}_0$ because there is no temporal sampling in this case. We also observe no skipping frames in $\mathcal{T}_1$ because this model only allows either pre-scanning or running the full pipeline. Each block in the table shows the results for a different temporal sampler $\mathcal{T}_M$, where $M$ determines the sampling range, \ie, the maximum number of frames to skip.

In general, the temporal samplers significantly reduce the average GFLOPS compared to $\mathcal{S}_0$, which is also scalable in terms of total TFLOPS. As the sampling range gets wider from $\mathcal{T}_1$ to $\mathcal{T}_4$, we can save more complexity while the accuracy gets more compromised. Furthermore, the speeding up factor is proportional to the sampling range $M$. We achieve the best accuracy with $\mathcal{S}_3, \mathcal{T}_1$, 0.5\% lower than $\mathcal{S}_0$ with 1.77x speed-up. On the other hand, $\mathcal{S}_0, \mathcal{T}_4$ has the highest speed-up of 4.01x but loses 5.71\% of accuracy. However, this model also provides the lowest trade-off factor between accuracy and speed-up. We see that the effect of using spatial sampler to compensate for the loss of accuracy is the most significant for $\mathcal{T}_1$, where more RoIs consistently result in higher top-1 accuracy. More RoIs also improves the speed-up factor of $\mathcal{T}_1$, compared against the corresponding spatial sampler counterparts. More interestingly, we observe that the percentage of pre-scanning frames has a low variance of 0.14\% regardless of the sampling range, while the amount of skipping frames increases when the such range raises. This behavior suggests that our temporal sampler knows how to compare the attention of frame $t$ with the hallucination from frame $t-1$, thus knows which frames only need to be pre-scanned. However, it is more challenging for the sampler to predict how many frames ahead to skip. We believe this happens because our hallucination is only predicted for only one future frame and will explore multi-frame hallucination in future work.

\section{In-detail figures}

\begin{figure}[h]
	\centering
	\includegraphics[width=0.9\linewidth]{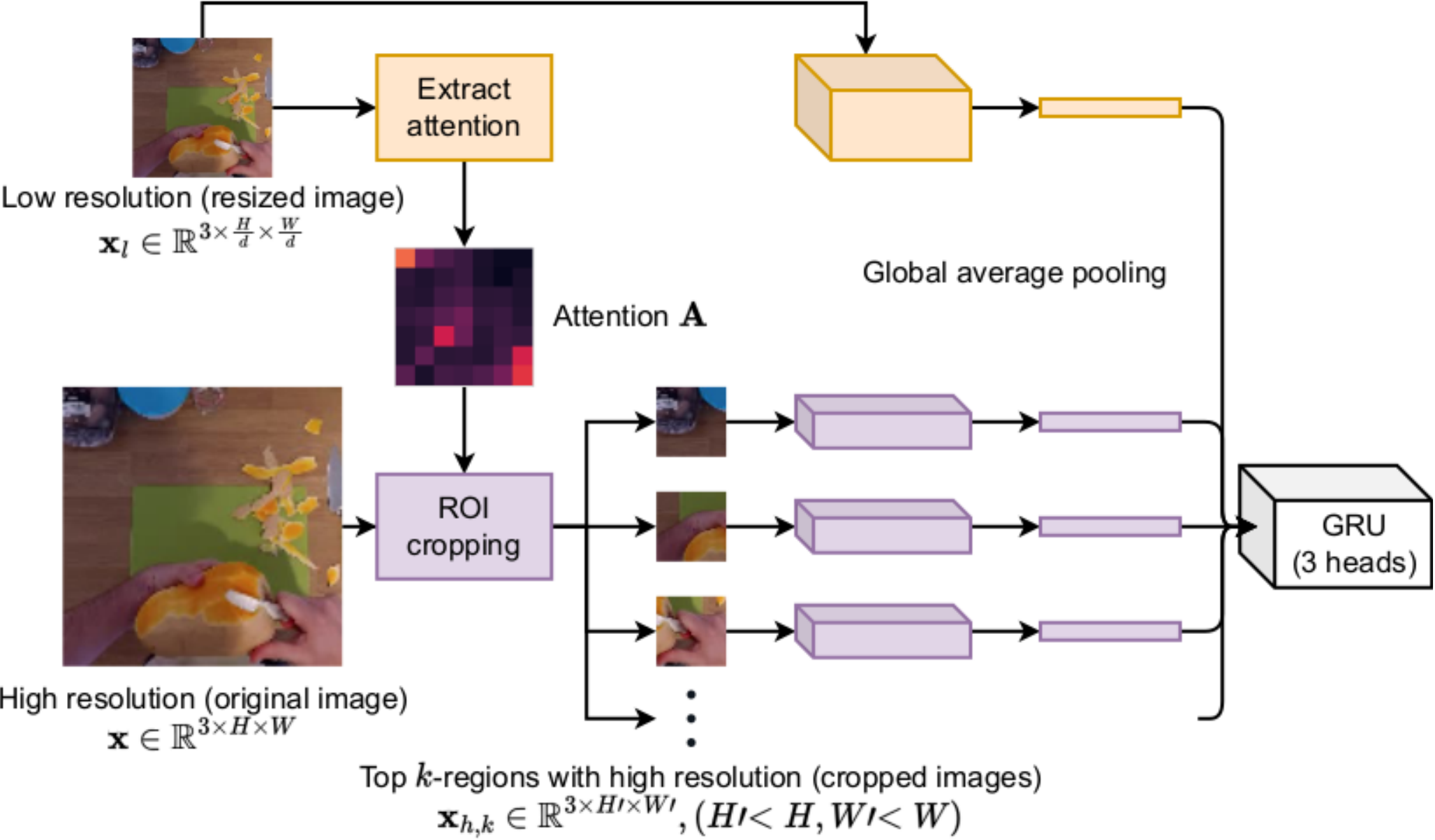}
	\caption{Spatial sampler uses attention from low-res image to sample the top-$k$ regions from the (original) high-res input. $\bx_l$ gives a global view, while $\bx_{h,k}$ provides local views at important regions of the original image $\bx$. The global average pooling at the end removes spatial dimension of the features, which are combined and fed to the GRU classifier.}
	\label{fig:space_sampler_big}
\end{figure}

\begin{figure}[h]
	\centering
	\includegraphics[width=0.9\linewidth]{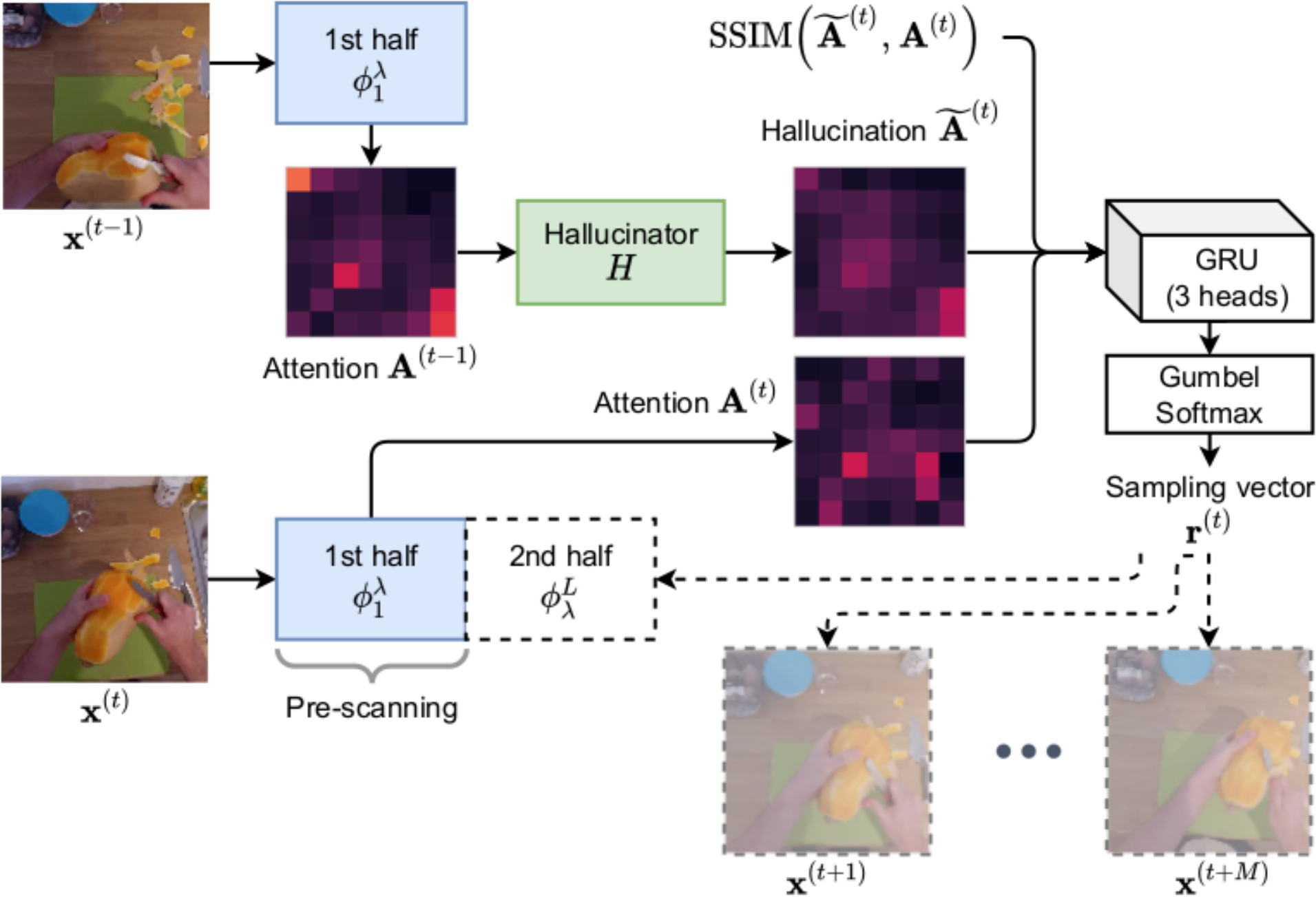}
	\caption{Temporal sampler with inputs at $t-1$ and $t$. Attention from the model's first half at time $t$, hallucination computed at time $t-1$, and their SSIM score are fed to a GRU to compute the sampling vector $\br^{(t)}$, deciding how many frames to skip (including the second half of the current frame). Model weights are shared across frames.}
	\label{fig:time_sampler_big}
\end{figure}

\begin{figure}[h]
    \centering
    \includegraphics[width=0.9\linewidth]{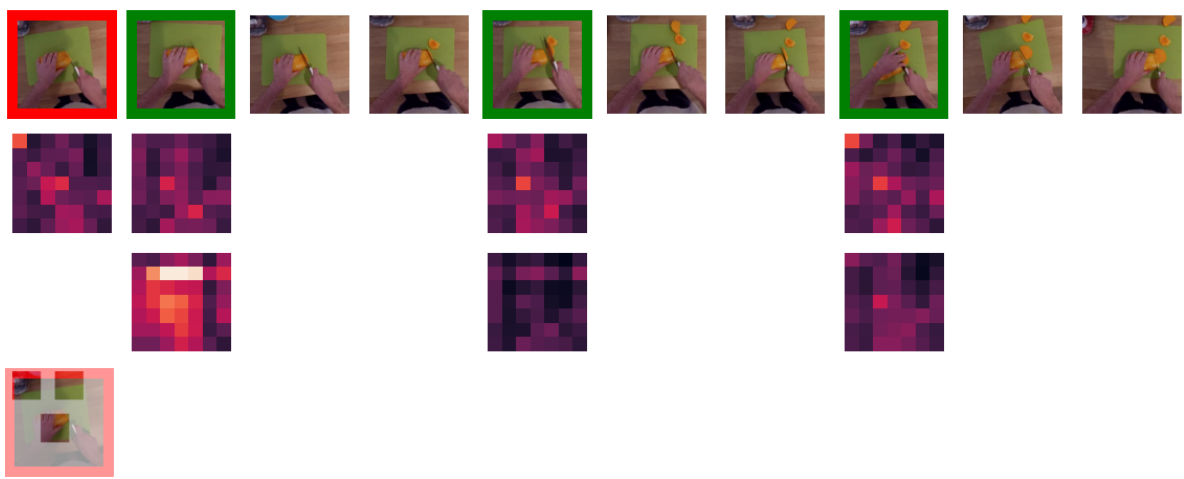}
    \caption{Qualitative results of action: cut squash}
    \label{fig:qualitative_qual87_big}
\end{figure}

\begin{figure}[h]
    \centering
    \includegraphics[width=0.9\linewidth]{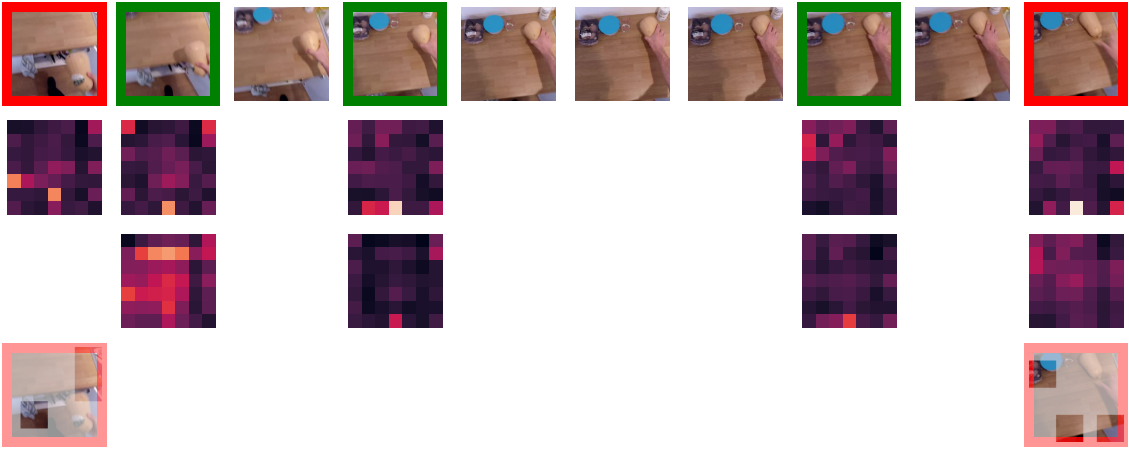}
    \caption{Qualitative results of action: put squash}
    \label{fig:qualitative_qual17_big}
\end{figure}

\begin{figure}[h]
    \centering
    \includegraphics[width=0.9\linewidth]{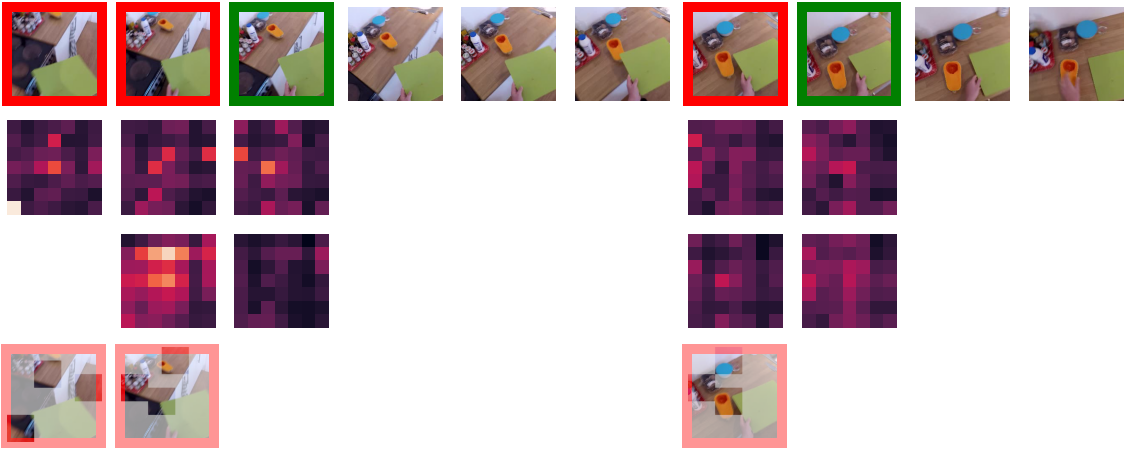}
    \caption{Qualitative results of action: open container}
    \label{fig:qualitative_qual57_big}
\end{figure}

\begin{figure}[h]
    \centering
    \includegraphics[width=0.9\linewidth]{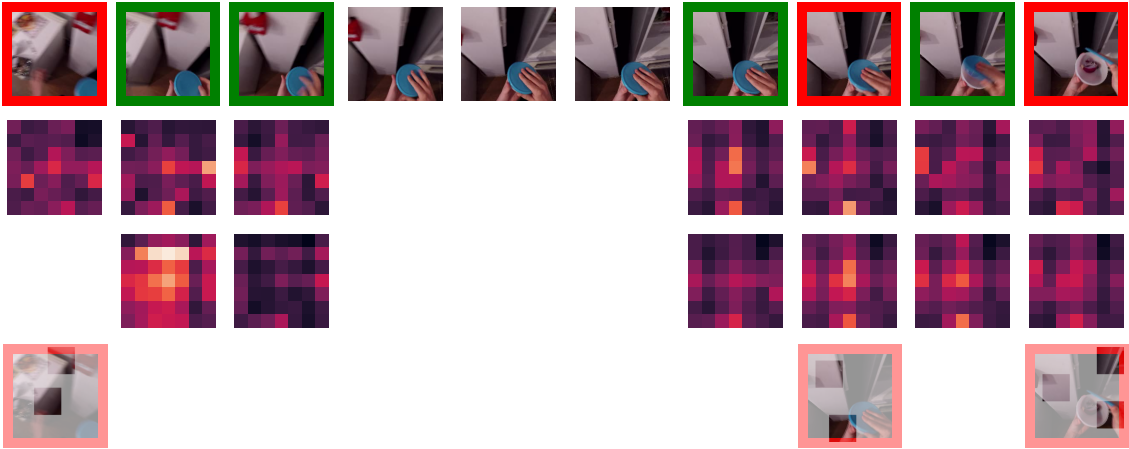}
    \caption{Qualitative results of action: open container}
    \label{fig:qualitative_qual3_big}
\end{figure}

This section provides figures with higher resolution and some minor modification for better visibility. Respectively, \figref{fig:space_sampler_big} and \figref{fig:time_sampler_big} correspond to \figref{fig:space_sampler} and \figref{fig:time_sampler} in the main paper. We also split the subfigures in \figref{fig:qualitative} into \figref{fig:qualitative_qual87_big}, \figref{fig:qualitative_qual17_big}, \figref{fig:qualitative_qual57_big}, and \figref{fig:qualitative_qual3_big} to increase the clarity.

\end{document}